\documentclass[10pt,journal,compsoc]{IEEEtran}
\usepackage{amsmath,amsfonts}
\usepackage{algorithmic}
\usepackage{algorithm}
\usepackage{array}
\usepackage{textcomp}
\usepackage{stfloats}
\usepackage{url}
\usepackage{verbatim}
\usepackage{graphicx}
\usepackage{cite}
\usepackage{multicol}
\usepackage{multirow}
\usepackage[table]{xcolor} 
\usepackage{xcolor}
\usepackage{booktabs}
\usepackage{makecell}
\usepackage{listings}
\usepackage{bm}
\usepackage{subfig}
\usepackage{tabularx}

\begin{document}

\title{Advancing Vision Transformer with Enhanced Spatial Priors}

\author{Qihang Fan, Huaibo Huang, Mingrui Chen, Hongmin Liu, Ran He, \textit{Fellow, IEEE}

\thanks{
Qihang Fan, Huaibo Huang, Mingrui Chen and Ran He are with the MAIS \& NLPR, Institude of Automation, Chinese Academic of Science, Beijing 100190, China.  (E-mail: fanqihang.159@gmail.com, huaibo.huang@cripac.ia.ac.cn, charmier2003@gmail.com, rhe@nlpr.ia.ac.cn)}
\thanks{ Hongmin Liu is with the School of Artificial Intelligence, University of Science and Technology Beijing, Beijing 100083, China. (E-mail: hmliu@ustb.edu.cn)
}
\thanks{Corresponding author: Hongmin Liu.}
}

\markboth{Journal of \LaTeX\ Class Files,~Vol.~14, No.~8, August~2021}%
{Shell \MakeLowercase{\textit{et al.}}: A Sample Article Using IEEEtran.cls for IEEE Journals}

\IEEEpubid{0000--0000/00\$00.00~\copyright~2021 IEEE}

\IEEEtitleabstractindextext{
\begin{abstract}
In recent years, the Vision Transformer (ViT) has garnered significant attention within the computer vision community. However, the core component of ViT, Self-Attention, lacks explicit spatial priors and suffers from quadratic computational complexity, limiting its applicability. To address these issues, we have proposed \textbf{RMT}, a robust vision backbone with explicit spatial priors for general purposes. RMT utilizes Manhattan distance decay to introduce spatial information and employs a horizontal and vertical decomposition attention method to model global information. Building on the strengths of RMT, \textbf{Euclidean enhanced Vision Transformer (EVT)} is an expanded version that incorporates several key improvements. Firstly, EVT uses a more reasonable Euclidean distance decay to enhance the modeling of spatial information, allowing for a more accurate representation of spatial relationships compared to the Manhattan distance used in RMT. Secondly, EVT abandons the decomposed attention mechanism featured in RMT and instead adopts a simpler spatially-independent grouping approach, providing the model with greater flexibility in controlling the number of tokens within each group. By addressing these modifications, EVT offers a more sophisticated and adaptable approach to incorporating spatial priors into the Self-Attention mechanism, thus overcoming some of the limitations associated with RMT and further enhancing its applicability in various computer vision tasks. Extensive experiments on Image Classification, Object Detection, Instance Segmentation, and Semantic Segmentation demonstrate that EVT exhibits exceptional performance. Without additional training data, EVT achieves \textbf{86.6\%} top1-acc on ImageNet-1k.
\end{abstract}
\begin{IEEEkeywords}
Vision Transformer, Spatial Prior, Token grouping.
\end{IEEEkeywords}}

\maketitle

\section{introduction}

The Vision Transformer (ViT)~\cite{vit} has become a highly-regarded visual architecture within the research community. Nonetheless, it encounters several significant issues. The core module of ViT, Self-Attention, inherently lacks explicit spatial priors, which is featured by the convolution. Moreover, the quadratic computational complexity of Self-Attention results in considerable computational costs when attempting to model global information, constraining its practical use.

Various studies have tried to mitigate these challenges~\cite{uniformer, SwinTransformer, CaiT, cloformer, cvt, cmt, LVT}. For example, the Swin Transformer~\cite{SwinTransformer} employs windowing operations to partition tokens for self-attention. This technique not only reduces the computational expense but also brings spatial priors into the model through the use of windows and relative position encoding. Likewise, NAT~\cite{NAT} alters the receptive field of Self-Attention to emulate the shape of convolutional layers, thereby decreasing computational costs and allowing the model to recognize spatial priors through its receptive field configuration. RMT~\cite{fan2023rmt} extends the concept of explicit decay in the NLP~\cite{retnet, alibi} to the spatial domain, devising a two-dimensional bidirectional spatial decay matrix based on the Manhattan distance between tokens. It also propose a horizontal and vertical decomposition attention mechanism to model the global information. 

In this work,  we also design a two-dimensional bidirectional spatial decay matrix based on the relative distance between tokens. In our spatial decay matrix, the attention scores for a target token decay more sharply for tokens that are further away. This design allows the target token to capture global information while also differentiating attention levels based on distance. By incorporating this spatial decay matrix, we were able to introduce explicit spatial priors into our vision backbone, enhancing its ability to handle spatial information effectively. Different from the Manhattan distance used in the RMT~\cite{fan2023rmt}, we use the Euclidean distance to model the relationship among tokens. This change is because human attention to objects far from the center of the visual field decays in a radial pattern~\cite{pervit}, which aligns with the radial growth pattern of Euclidean distance.  Besides, we adopt a one-dimensional, spatially-independent token grouping method. Compared to two-dimensional, spatially-dependent grouping methods, such as the windowing in Swin Transformer and the two-dimensional dilated grouping in MaxViT, this approach allows for more flexible control over the number of tokens within each group. Our experiments demonstrate that the the proposed spatial decay matrix brings strong spatial prior to the model. The 1D token grouping method, combined with the spatial decay matrix, can achieve better results than the 2D grouping method. Since our model incorporates spatial priors by utilizing the Euclidean distance between tokens, we name it the \textbf{Euclidean Enhanced Vision Transformer (EVT)}.



\begin{figure*}[ht]
    \centering
    \includegraphics[width=0.8\linewidth]{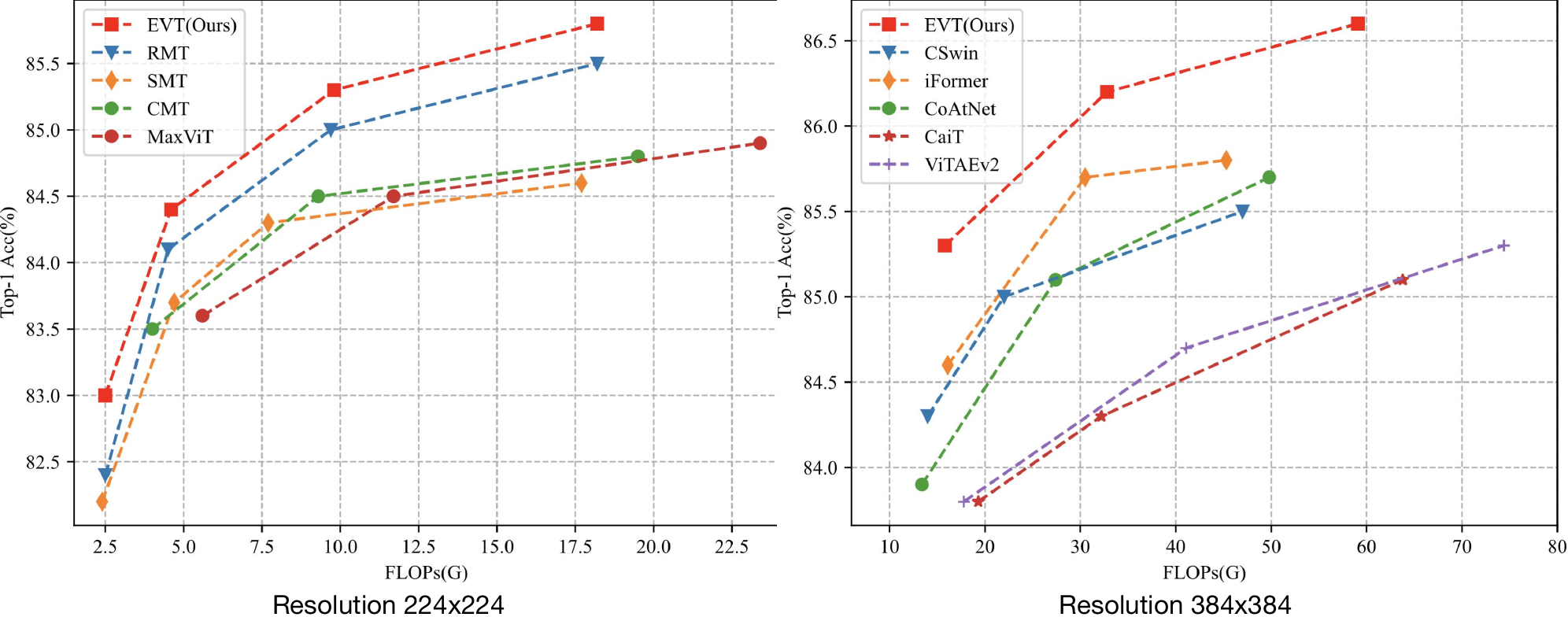}
    \caption{FLOPs v.s. Top-1 accuracy on ImageNet-1K.}
    \label{fig:flops-acc}
\end{figure*}


We conduct extensive experiments, such as image classification, object detection, instance segmentation, semantic segmentation, and robustness tests, to validate the performance of EVT. EVT demonstrates strong performance advantages across all tasks. As shown in Fig.~\ref{fig:flops-acc}, at a resolution of $224\times 224$, EVT achieves a top-1 accuracy of \textbf{85.8\%} without relying on any additional training data or supervision information, while requiring only \textbf{18.2 GFLOPs} of computational effort. When the resolution is increased to $384\times 384$, the model's performance is further enhanced. Our EVT-L, with just 100M parameters, achieves a top-1 accuracy of \textbf{86.6\%}, significantly surpassing existing models. 

A preliminary version of this work was published in the CVPR 2024~\cite{fan2023rmt}. In this paper, we extend our conference version in the following aspects:

\begin{itemize}
    \item We replace the Manhattan distance-based spatial priors in RMT with the more intuitive Euclidean distance-based spatial priors. We conduct extensive experiments to demonstrate that this radially decaying spatial prior significantly enhances the model's spatial understanding, thereby improving its performance.
    \item  We propose a one-dimensional grouping method to group vison tokens. This grouping method ignoring the spatial relationships between tokens, allowing the model to flexibly control the number of tokens within each group. Compared to the horizontal and vertical decomposition attention mechanism used in the RMT, the proposed grouping method is simpler and faster.
    \item We conduct extensive experiments to validate the performance of EVT. EVT demonstrates exceptional performance across various tasks, including image classification, object detection, instance segmentation, semantic segmentation, and OOD dataset classification. Additionally, we perform numerous ablation studies to verify the contribution of each module within EVT.
\end{itemize}
\section{Related works}
\subsection{Vision Transformers}
Since the introduction of the original plain ViT~\cite{vit}, many works have focused on designing hierarchical architectures that can better capture spatial information and multi-scale features~\cite{SwinTransformer, cswin, dat, davit, biformer,convnext, cmt, fan2024semantic, fan2024vision}. The core of these works primarily revolves around designing efficient, linear-complexity attention mechanisms that can more effectively provide inductive biases. For example, Swin Transformer introduced window-based self-attention~\cite{SwinTransformer, twins}, PVT/PVTv2 proposed spatially down-sampled attention~\cite{pvt, pvtv2, cmt, FAT}, DAT presented deformable attention~\cite{dat}, and BiFormer developed multi-scale routing attention~\cite{biformer}. There are also several approaches that utilize global or region tokens to convey information between different areas of the image~\cite{ViL, regionvit, vip, perceiver, msgtransformer}. Additionally, many works have attempted to scale ViT, approaching the problem from both resolution and model parameter perspectives~\cite{HRFormer, HRViT, sis, hirivit, swinv2, internimage}. Apart from the aforementioned methods, many other approaches aim to accelerate ViT inference. Among these, methods like EViT~\cite{evit, dynamicvit} employ token pruning, while ToMe~\cite{tome} uses token merging to combine similar tokens. Additionally, some approaches attempt to reduce the computation overhead of global attention by clustering tokens~\cite{stvit,fan2024semantic}. In this work, we introduce a novel spatially-independent one-dimensional token grouping method. This approach allows for more flexible control over the number of tokens within each group. With the support of spatial priors, this method has demonstrated impressive performance.

\subsection{Convolution-Transformer Hybrid Architectures.}
Convolutions are effective at capturing high-frequency texture information in images, whereas Transformers excel at modeling low-frequency global information~\cite{conformer, cloformer, cmt}. Consequently, many architectures have attempted to combine the strengths of both approaches~\cite{mixformer, iformer, cmt, litv2, uniformer, conformer, coatnet, cloformer}. CMT~\cite{cmt} combines lightweight attention modules with convolutions, sequentially extracting both local and global information. iFormer~\cite{iformer} adopts a parallel design, where within the same token mixer module, one portion of the channels employs self-attention mechanisms to extract global information, while another portion uses convolutions to capture local information. These pieces of information are then combined using linear projection. Moreover, many other works insert lightweight convolutions into various subcomponents of the Transformer design. For example, using a Conv Stem at the beginning of the model for downsampling the image~\cite{stvit, biformer}, inserting convolution-based local information enhancement modules before the Attention mechanism~\cite{cmt, stvit, uniformer}, and introducing positional information in the FFN with convolutions~\cite{pvt, pvtv2,Ortho}. In the design of both RMT~\cite{fan2023rmt} and EVT, convolutions are also used to enhance the model's local representation capabilities.

\begin{figure*}[t]
    \centering
    \includegraphics[width=0.8\linewidth]{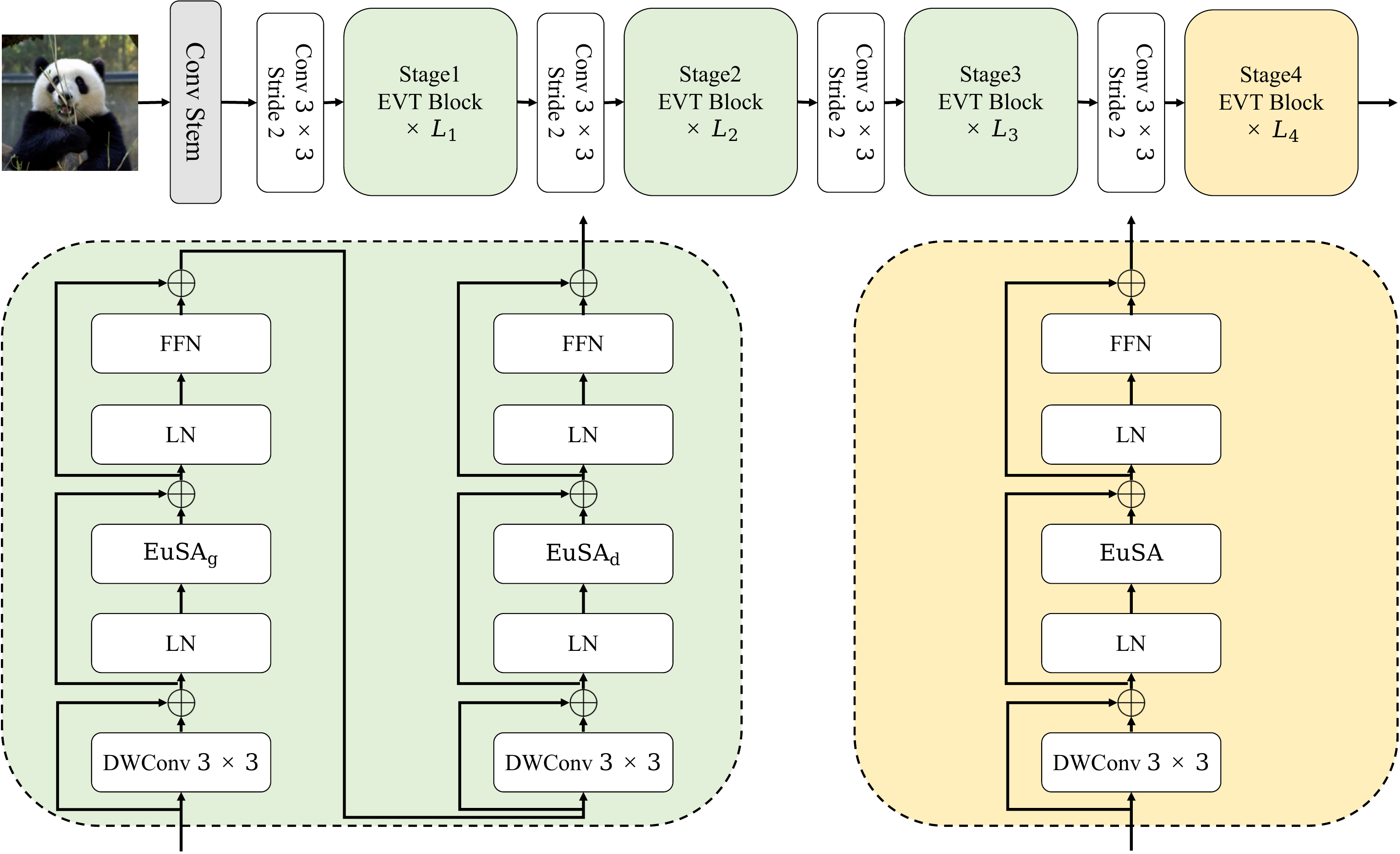}
    \caption{Overall Architecture of EVT.}
    \label{fig:main}
\end{figure*}

\subsection{Position Prior in Vision Models}
Positional encoding is a crucial module for Transformers as it imparts positional information to each token, thereby enabling the Transformer to perceive the positions of tokens~\cite{attention}. The earliest ViT utilized sinusoidal-based absolute positional encoding~\cite{vit}. Subsequently, many works have sought to improve the positional encoding of visual tokens~\cite{CPVT, cswin, swinv2, SwinTransformer, pvtv2, EVA02}. CPVT~\cite{CPVT} introduced Conditional Positional Encoding (CPE) based on depthwise separable convolutions, which can be very flexibly inserted at any position within a ViT. In Swin Transformer v2~\cite{swinv2}, log-spaced relative positional encoding was proposed. Compared to the original relative positional encoding, this approach is better suited for high-resolution images~\cite{SwinTransformer}. In CSwin~\cite{cswin}, LePE is employed, which is a highly flexible convolution-based positional encoding method. This approach has been widely adopted in many techniques~\cite{han2023flatten, han2023agent, biformer}. Additionally, some methods incorporate convolution into the FFN (Feed-Forward Network) module to provide positional information, thereby enhancing the performance of Transformers~\cite{pvtv2, cmt, Ortho}. Unlike previous methods, EVA02~\cite{EVA02} drew inspiration from the widely-used rotary positional encoding (RoPE) in large language models~\cite{su2021roformer}, proposing a two-dimensional rotary positional encoding and applying it to vision models. In RMT~\cite{fan2023rmt}, inspired by the success of ALiBi~\cite{alibi} and RetNet~\cite{retnet} in the NLP domain, we introduced Manhattan distance-based explicit spatial decay into the self-attention mechanism to provide the model with spatial information. In this work, we further improved this spatial decay by replacing the Manhattan distance with the more intuitive Euclidean distance.

\section{method}
\subsection{preliminary: RMT}
In RMT, we drew inspiration from the one-dimensional temporal decay proposed in RetNet~\cite{retnet} and ALiBi~\cite{alibi}, and extended it to the two-dimensional spatial domain. This resulted in the Manhattan distance-based spatial decay used in RMT. As the relative distance between tokens increases, this decay progressively intensifies, thereby introducing spatial priors into the model.

Specifically, as shown in Eq.~\ref{eq:MaSA}, our Manhattan self-attention mechanism incorporates a $D$ matrix as a decay factor in the attention matrix, thereby introducing spatial information into the self-attention mechanism:
\begin{equation}
    \label{eq:MaSA}
    \begin{aligned}
        \mathrm{MaSA}(X) &= (\mathrm{Softmax}(Q K^T )\odot D^{2d})V \\
        D_{nm}^{2d}&=\gamma^{|x_n-x_m|+|y_n-y_m|}
    \end{aligned}
\end{equation}
Here, $(x_n, y_n)$ represent the horizontal and vertical coordinates of token $n$.

Additionally, hierarchical architectures tend to have high resolutions in the shallow layers, leading to significant computational overhead. To address this challenge, RMT introduces a decomposed form of the self-attention mechanism, which allows for global information modeling with lower computational cost. Specifically, it calculates attention weights using one-dimensional decay matrices along the horizontal and vertical directions of the image, then applies these attention weights to the value. The detailed process is shown in Eq.~\ref{eq:attnscore}:
\begin{equation}
    \label{eq:attnscore}
    \begin{aligned}
        &Attn_H=\mathrm{Softmax}(Q_H K_H^T) \odot D^{H},\\
        &Attn_W=\mathrm{Softmax}(Q_W K_W^T) \odot D^{W}, \\
        &\mathrm{MaSA}(X)=Attn_H(Attn_WV)^T; \\
    \end{aligned}
\end{equation}

\subsection{Overall Architecture of EVT}

The overall structure of EVT is depicted in Fig.~\ref{fig:main}. For an input image $x \in \mathbb{R}^{3 \times H \times W}$, we initially downsample it using a Conv Stem. This results in tokens having a shape of $C_1 \times \frac{H}{4} \times \frac{W}{4}$. Inspired by previous works~\cite{SwinTransformer, pvt, cswin}, we employ a hierarchical configuration to process these tokens. The model is organized into four stages corresponding to downsampling factors of $\frac{1}{4}$, $\frac{1}{8}$, $\frac{1}{16}$, and $\frac{1}{32}$, respectively. This hierarchical setup is advantageous for downstream tasks like object detection, facilitating the construction of feature pyramids.

An EVT block comprises three main modules: Conditional Positional Encoding (CPE)~\cite{CPVT}, Euclidean Self-Attention (EuSA), and a classical Feed-Forward Network (FFN)~\cite{attention}. A full EVT block can be expressed by Eq.~\ref{eq:block}:

\begin{equation}
\label{eq:block}
\centering
\begin{split}
    &X={\rm CPE(} X_{in} {\rm )} + X_{in}, \\
    &Y={\rm EuSA(} {\rm LN(} X {\rm ))} + X, \\
    &Z={\rm FFN(} {\rm LN(} Y {\rm ))} + Y.
\end{split}
\end{equation}

In each block, the input tensor $X_{in} \in \mathbb{R}^{C \times H \times W}$ first goes through the CPE module to introduce positional information. The EuSA module then acts as the token mixer. Finally, the FFN integrates the channel-wise information of the tokens.

\subsection{Manhattan Distance to Euclidean Distance}

In RMT, the explicit spatial decay is based on the Manhattan distance, which does not align with the way the human eye perceives images. When recognizing images, human attention decays radially relative to distance~\cite{pervit}. So we replaced the Manhattan distance with the Euclidean distance. We analyze the advantages of Euclidean distance over Manhattan distance from two perspectives.

\textbf{(1) From the Perspective of Distribution Similarity: }The core purpose of using a decay matrix is to incorporate a spatially relevant prior into the attention scores. While standard self-attention can also learn this prior, our decay matrix explicitly embeds it into the attention scores, thereby simplifying the learning process. The closer the attention score distribution of a well-trained ViT model without a decay matrix is to that of our decay matrix, the more effectively our decay matrix facilitates the learning of spatial priors, ultimately leading to improved model performance. 

Based on this consideration, we train an EVT-T model without any decay matrix. Then, we analyze the correlation between its attention score distribution and those of different decay matrices. We use Jensen-Shannon (JS) divergence to measure the similarity between different distributions, which is computed as follows:  

\begin{equation}
    D_{\text{JS}}(P || Q) = \frac{1}{2} D_{\text{KL}}(P || M) + \frac{1}{2} D_{\text{KL}}(Q || M)
\end{equation}

where \( M = \frac{1}{2} (P + Q) \) is the average distribution, and \( D_{\text{KL}}(P || Q) \) denotes the Kullback-Leibler (KL) divergence, defined as  
\begin{equation}
    D_{\text{KL}}(P || Q) = \sum_{i} P(i) \log \frac{P(i)}{Q(i)}.
\end{equation}

a smaller JS divergence value indicates a higher correlation between the distributions. We compute the average JS divergence for the images in the ImageNet-1K validation set. The results are shown in the Tab.~\ref{tab:js}. Based on the JS divergence values, the distribution of the Euclidean distance-based decay matrix exhibits higher similarity to the distribution of the standard attention scores in the trained model, which also leads to better performance.
\begin{table}[h]
    \centering
    \begin{tabular}{c|c c}
    \toprule[1pt]
    \multicolumn{3}{c}{Comparison on EVT-T}\\
    \midrule[0.5pt]
     & JS$\downarrow$ & Acc(\%) \\
    no decay matrix & -- & 82.3 \\
    Mahattan distance & 0.20 & 82.7(\textcolor{red}{+0.4}) \\
    Euclidean distance & 0.12 & 83.0(\textcolor{red}{+0.7}) \\
    \bottomrule[1pt]
    \end{tabular}
    \caption{The JS divergence between different decay matrices and the attention scores of the model without a decay matrix, and the performance of models using different decay matrices. }
    \label{tab:js}
\end{table}

{The distribution of our decay matrix closely resembles the attention score distribution of a well-trained ViT model. The latter already contains the spatial knowledge that the native ViT learns, although this spatial prior remains relatively weak due to the absence of external intervention. Because the two distributions are highly similar, their combination further strengthens the spatial priors learned by the model, thereby leading to improved model performance.} 

\begin{table}[h]
    \centering
    \begin{tabular}{c|c c c}
    \toprule[1pt]
    Model & JS$\downarrow$ & Acc(\%) & mIoU(\%) \\
    \midrule[0.5pt]
    DeiT-B & 0.24 & 81.8 & 17.8 \\
    EVT-DeiT-B & 0.10 & 82.3(\textcolor{red}{+0.5})& 24.6(\textcolor{red}{+6.8}) \\
    \midrule[0.5pt]
    DINOv2-ViT-B & -- & 84.5 & 47.3 \\
    \bottomrule[1pt]
    \end{tabular}
    \caption{The JS divergence between different models. }
    \label{tab:js}
\end{table}

{To further illustrate our assertion that the spatial decay matrix can enhance the spatial priors of the native ViT, we compare different models with DINOv2—a powerful ViT backbone that undergoes large-scale unsupervised training on extensive datasets. The results are shown in Tab.~\ref{tab:js}. Compared to the original DeiT, our EVT-DeiT exhibits an attention distribution that is more similar to that of DINOv2, indicating that EVT-DeiT acquires richer spatial knowledge. We pretrain the models on ImageNet-1K and conduct linear probing on ADE20K using the pretrained models. EVT-DeiT achieves significantly better results than DeiT-T, which demonstrates that it learns more spatial knowledge.}

\textbf{(2) From the Perspective of Numerical Stability:} The Manhattan distance is defined as:
\begin{equation}
    d_{\text{Manhattan}}(x, y) = |x_1 - y_1| + |x_2 - y_2|
\end{equation}
While straightforward, it presents \textbf{Non-Smooth Distance Variation:} The Manhattan distance increases linearly along coordinate axes but remains constant along diagonals. This results in an uneven scaling of the spatial decay factor, potentially introducing anisotropic biases in attention modulation.

{\textbf{(3) Comparison with Standard Attention Mechanism}
For standard self-attention (without spatial decay), the attention weights are:
\begin{equation}
    A^{\text{std}}_{nm} = \frac{\exp(Q_n K_m^T/\sqrt{d})}{\sum_{k} \exp(Q_n K_k^T/\sqrt{d})}
\end{equation}
This formulation does not explicitly encode any spatial prior, treating all tokens equally regardless of their geometric positions. While standard attention mechanisms are capable of learning spatial relationships, they typically require very large-scale training data and extensive optimization to capture meaningful spatial structures (as seen in models such as DINOv2). Without an explicit spatial bias, the learned spatial features tend to be weaker and less robust, especially when training data or computational resources are limited.}

{By introducing the spatial decay matrix, the attention weights become:
\begin{equation}
    A_{nm} = \frac{\exp(Q_n K_m^T/\sqrt{d}) \cdot E_{nm}}{\sum_{k \in \mathcal{N}_n} \exp(Q_n K_k^T/\sqrt{d}) \cdot E_{nk}}
\end{equation}
where $E_{nm}$ can be either L2 or L1-based.}

{\textbf{(4) Gradient of Attention Weights with Respect to Spatial Coordinates}
For L2-based decay:
\begin{equation}
    E^{2d}_{nm} = \gamma^{\sqrt{(x_n - x_m)^2 + (y_n - y_m)^2}}
\end{equation}
\begin{equation}
    \frac{\partial A_{nm}}{\partial x_n} = \log \gamma \cdot A_{nm} \left( \frac{x_n - x_m}{d_{nm}} - \sum_{k \in \mathcal{N}_n} A_{nk} \frac{x_n - x_k}{d_{nk}} \right)
\end{equation}
where $d_{nm} = \sqrt{(x_n - x_m)^2 + (y_n - y_m)^2}$.}

{For L1-based decay:
\begin{equation}
    E^{1d}_{nm} = \gamma^{|x_n - x_m| + |y_n - y_m|}
\end{equation}
\begin{equation}
    \begin{split}
        \frac{\partial A_{nm}}{\partial x_n} = \log \gamma \cdot A_{nm} \Big( \text{sign}(x_n - x_m) \\
        \qquad - \sum_{k \in \mathcal{N}_n} A_{nk} \text{sign}(x_n - x_k) \Big)
    \end{split}
\end{equation}
The L2 gradient is smooth and direction-aware, while the L1 gradient is piecewise constant and discontinuous, which can hinder optimization and spatial generalization.}

{\textbf{(5) Spectral Analysis and Spatial Coverage}
The spatial decay matrix $E_{nm}$ can be interpreted as a weighted adjacency matrix of a graph over tokens. The Laplacian matrix $L = D - E$ (where $D$ is the degree matrix) encodes the spatial connectivity. For L2 decay, the eigenvectors of $L$ capture isotropic spatial harmonics, allowing the model to represent spatial features smoothly across all directions. In contrast, L1 decay leads to axis-aligned harmonics, limiting the diversity of spatial patterns.}

{\textbf{(6) Information-Theoretic Perspective: Spatial Entropy Optimization}
The spatial decay acts as a prior that shapes the entropy of the attention distribution:
\begin{equation}
    H(A_n) = -\sum_{m \in \mathcal{N}_n} A_{nm} \log A_{nm}
\end{equation}
L2 decay maximizes entropy under isotropic constraints, encouraging the model to capture diverse and meaningful spatial dependencies. L1 decay, due to its axis-aligned bias, can reduce entropy and restrict the model's ability to learn complex spatial relationships.}

{\textbf{(7) Expressivity and Generalization}
The output for each token is:
\begin{equation}
    Y_n = \sum_{m \in \mathcal{N}_n} A_{nm} V_m
\end{equation}
With L2 decay, the effective receptive field is:
\begin{equation}
    \text{ReceptiveField}_n = \{ m \mid A_{nm} > \epsilon \}
\end{equation}
which adapts smoothly to spatial structure. L1 decay tends to produce axis-aligned receptive fields, limiting expressivity in complex spatial scenarios.}

{\textbf{(8) Gradient Flow and Optimization Landscape}
The second derivative for L2 decay is:
\begin{equation}
    \frac{\partial^2 E^{2d}_{nm}}{\partial x_n^2} = E^{2d}_{nm} \cdot (\log \gamma)^2 \cdot \frac{(y_n - y_m)^2}{((x_n - x_m)^2 + (y_n - y_m)^2)^{3/2}}
\end{equation}
This ensures a well-conditioned optimization landscape, supporting stable and efficient gradient flow. In contrast, L1 decay produces zero second derivatives almost everywhere, except at discontinuities, which can lead to training instability.}

{\textbf{(9) Unified View: Why L2 is Superior for Spatial Feature Learning}
While both L1 and L2 decays can enforce locality, L2's rotational invariance and smooth decay better match the natural geometry of images and spatial patterns. This is particularly important in grouped and dilated attention, where token neighborhoods are irregular. L2 decay enables the model to learn spatial priors adaptively from all directions, resulting in more coherent aggregation, better generalization, and improved representation of complex spatial structures. Even in standard attention architectures, L2 decay enhances spatial expressivity and optimization, outperforming both L1 and no-decay baselines.}

Based on above analysis, we employ the Euclidean distance:
\begin{equation}
    d_{\text{Euclidean}}(x, y) = \sqrt{(x_1 - y_1)^2 + (x_2 - y_2)^2}
\end{equation}
which offers \textbf{Smooth and Continuous Scaling:} Unlike the step-wise behavior of the Manhattan distance, the Euclidean distance varies smoothly, ensuring a gradual and consistent decay of attention scores. The smooth variation of Euclidean distance leads to a well-behaved weighting function, preventing sudden changes in attention distribution and ensuring stable training.

Since our method applies a distance-based decay function to attention scores, the discontinuous nature of the Manhattan distance can cause instability in attention modulation. In contrast, the Euclidean distance, with its smooth transitions, ensures a more stable and consistent weighting mechanism, ultimately improving attention performance. Based on these, we believe that using Euclidean distance as the decay factor for tokens should be a more intuitive approach. We have upgraded the Manhattan Self-Attention mechanism (MaSA) in Eq.~\ref{eq:MaSA} to the Euclidean Self-Attention mechanism (EuSA) based on Euclidean distance, as detailed in Eq.~\ref{eq:EuSA}:
\begin{equation}
    \label{eq:EuSA}
    \begin{aligned}
        \mathrm{EuSA}(X) &= (\mathrm{Softmax}(Q K^T )\odot E^{2d})V \\
        E_{nm}^{2d}&=\gamma^{\sqrt{(x_n-x_m)^2+(y_n-y_m)^2}}
    \end{aligned}
\end{equation}
Here, $E$ is the decay matrix based on Euclidean distance, where each element $E_{mn}$ represents the Euclidean distance between token $m$ and token $n$. Moreover, $\gamma$ is a manually set hyper-parameter, and its value differs for each head in the multi-head self-attention mechanism. The value of $\gamma$ controls the receptive field of the tokens, and using different $\gamma$ values for different heads enables the model to perceive multi-scale features.

\begin{figure*}
    \centering
    \includegraphics[width=0.8\linewidth]{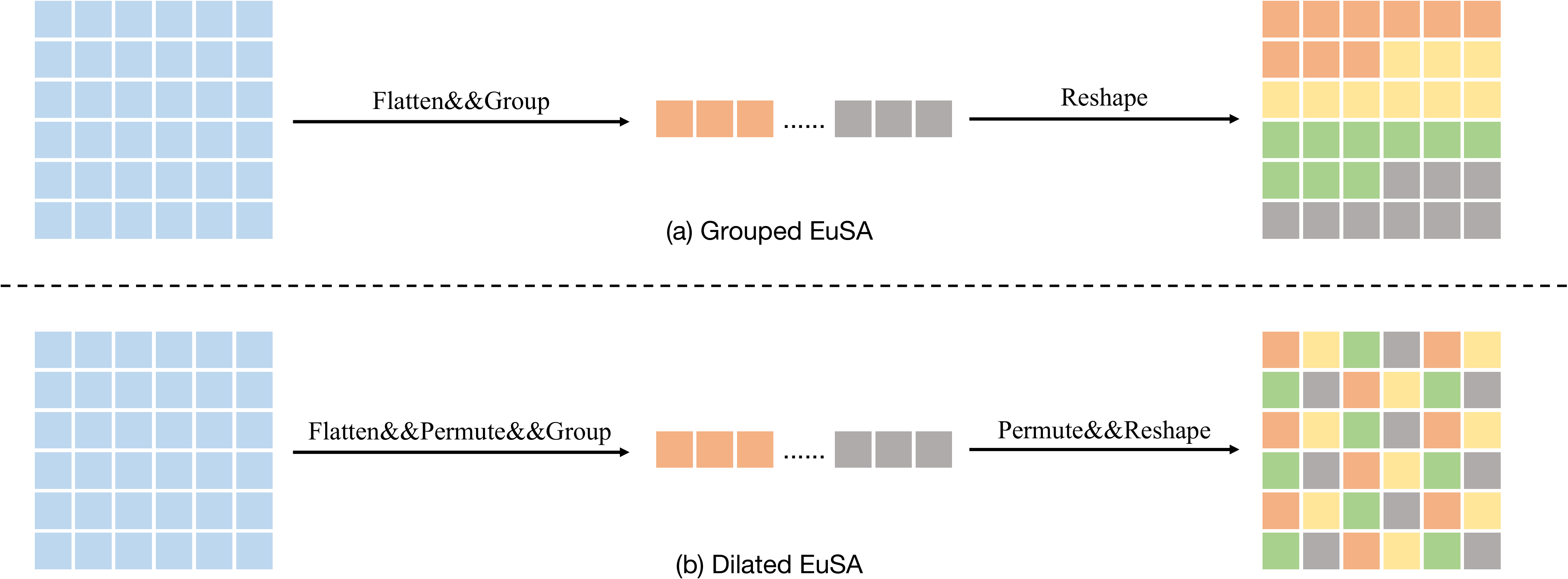}
    \caption{Illustration of Grouped EuSA (${\rm EuSA_g}$) and Dilated EuSA (${\rm EuSA_d}$). Different colors represent different groups.}
    \label{fig:group}
\end{figure*}

\subsection{Decomposed Form to the Grouped Form}

In RMT, we decompose global attention into horizontal and vertical dimensions and then weight the values using the attention weights from each dimension, allowing MaSA to perceive global information, as shown in Eq.~\ref{eq:attnscore}. However, this approach presents two issues. 

First, this method results in higher complexity. Compared to other linear complexity attention mechanisms, such as Window Self-Attention~\cite{SwinTransformer}, the complexity of MaSA is:
\begin{equation}
    \label{eq:complex}
    \begin{aligned}
        O(hwwd)+O(hhwd)\geq O(N^{1.5}d).
    \end{aligned}
\end{equation}
Here, $h$ and $w$ are the height and width of the feature map, respectively, and $N=hw$. The equality in Eq.~\ref{eq:complex} holds if and only if $h = w = \sqrt{N}$. The high complexity limits the model's performance in high-resolution tasks, such as object detection and semantic segmentation.

Second, the decomposition approach of MaSA self-attention hinders the model's parallelism. The horizontal and vertical attention scores cannot be used to weight the values in parallel. Their multiplication with $V$ follows a strict sequential order: $Attn_w $ is first multiplied with $V$, and then $Attn_h$ is applied to the resulting product. This step significantly limits the parallelism of the model and is impossible to optimize through engineering techniques. Consequently, although RMT has a low computational complexity, its inference speed remains relatively slow.
\textcolor{red}{
\begin{figure}
    \centering
    \includegraphics[width=0.8\linewidth]{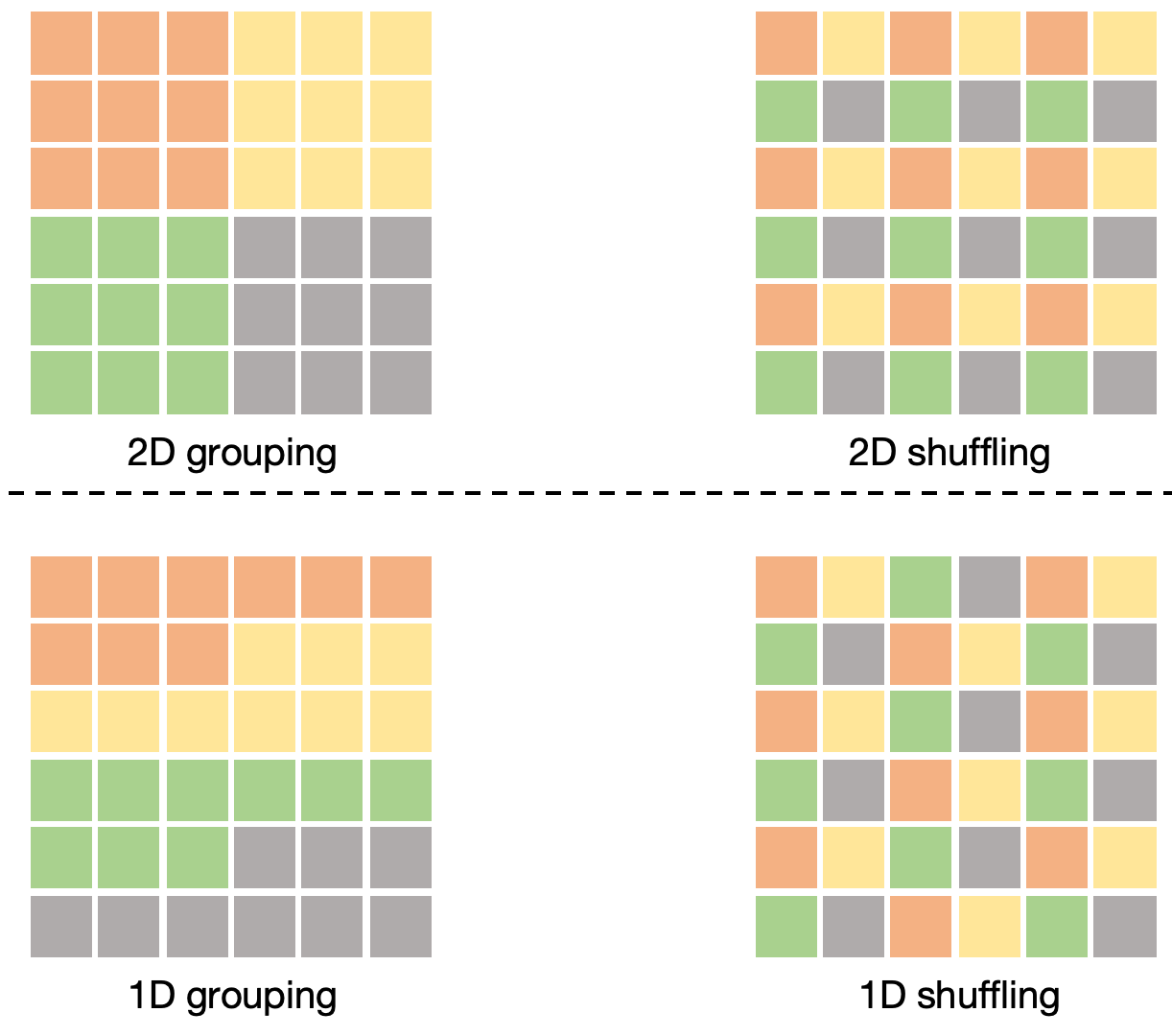}
    \caption{Comparison between 2D grouping/shuffling and our 1D grouping/shuffling.}
    \label{fig:grsh}
\end{figure}}

\begin{table}[ht]
    \centering     
    \setlength{\tabcolsep}{0.8mm}
    \begin{tabular}{c|c|c|ccc|cc}
    \toprule[1pt]
         Group & n\_tokens & EuSA & \makecell{FLOPs\\(G)} & \makecell{Throughput\\(imgs/s)} & \makecell{acc\\(\%)} & \makecell{FPS\\(imgs/s)} & \makecell{mIoU\\(\%)} \\
         \midrule[0.5pt]
         2d & $7\times7$ & $\times$ & 4.4 & 1002 & 83.7 & 21.6 & 49.2 \\
         2d & $7\times7$ & $\checkmark$ & 4.4 & 926 & 83.8 & 21.5 & 49.3 \\
         2d & $8\times8$ & $\checkmark$ & 4.6 & 889 & 83.9 & 21.0 & 49.4 \\
         2d & $10\times10$ & $\checkmark$ & 4.8 & 806 & 84.1 & 20.4& 49.7 \\
         \midrule[0.5pt]
         1d & 49 & $\times$ & 4.4 & 1168 & 83.5 & 23.2 & 48.8 \\
         1d & 49 & $\checkmark$ & 4.4 & 1102 & 83.9 & 23.1 & 49.6 \\
         1d & 64 & $\checkmark$ & 4.5 & 1052 & 84.2 & 22.8 & 49.8 \\
         1d & 98 & $\checkmark$ & 4.6 & 1001 & 84.4 & 22.4 & 50.0 \\
         \bottomrule[1pt]
    \end{tabular}
   \caption{Ablation study on different grouping/shuffling strategy. "n\_token" represents the number of tokens within each group/grid, and "EuSA" indicates whether the Euclidean decay matrix is used.}
    \label{tab:abgroup}
\end{table}
\begin{table}[ht]
    \centering
    \begin{tabular}{c|c c c}
    \toprule[1pt]
         Grouping/Shuffling & 224 & 384 & 512 \\
         \midrule
         1D & 83.0 & 83.3/84.0 & 82.8/84.5 \\
         2D & 82.8 & 82.6/83.6 & 81.1/84.0 \\
         \bottomrule[1pt]
    \end{tabular}
    \caption{Comparison of different grouping/shuffling methods. The results on the left show the direct inference of the model trained at a resolution of 224x224. The results on the right are from the model after fine-tuning at a higher resolution.}
    \label{tab:res}
\end{table} 
Based on the two considerations above, we abandoned the decomposition approach of global self-attention in MaSA and instead adopted a token grouping method to reduce the computational cost of the model. Most token grouping methods, whether it is window grouping or dilated window grouping, consider the two-dimensional spatial characteristics of tokens, i.e., grouping tokens simultaneously in both the height and width dimensions~\cite{SwinTransformer, cswin, maxvit}. Although this introduces inductive bias to ViT, such a grouping method lacks flexibility, as shown in Tab.~\ref{tab:abgroup}, the 2D grouping method has a relatively large impact on the model's throughput. Therefore, we adopted a one-dimensional token grouping approach, where the feature map is flattened before grouping the tokens. Given that our Euclidean decay matrix has already introduced sufficient spatial priors to the model, the one-dimensional grouping method does not degrade the model's performance. To ensure that the model has a global receptive field, we propose two token grouping methods: grouped EuSA (${\rm EuSA_g}$) and dilated EuSA (${\rm EuSA_d}$). The specific grouping methods are illustrated in Fig.~\ref{fig:group}. The comparison between the two approaches is shown in the Fig.~\ref{fig:grsh}. In the first three stages of the model, we alternated between ${\rm EuSA_g}$ and ${\rm EuSA_d}$ to ensure that the model has a global receptive field. In the final stage, due to the smaller number of tokens, we used the original EuSA.

{Compared to previous 2D Group/Dilated Attention, the proposed 1D Group/Dilated Attention has following advantages.}

{\textbf{(1) Theoretical Comparison: Receptive Field of 1D vs. 2D Group/Dilated Attention}
Let $N = H \times W$ be the total number of tokens, $w$ the number of tokens each group attends to per layer (after dilation), $g$ the number of groups, and $L$ the number of layers. We ensure that in both 1D and 2D group/dilated attention, each group attends to exactly $w$ tokens per layer.}

{\textbf{Initial receptive field:}  
At layer $l=1$, for any token $n$:
\begin{equation}
    R_n^{(1)} = \mathcal{N}_n
\end{equation}
where $|\mathcal{N}_n| = w$.}

{\textbf{Recursive expansion:}  
At layer $l+1$:
\begin{equation}
    R_n^{(l+1)} = R_n^{(l)} \cup \left( \bigcup_{m \in R_n^{(l)}} \mathcal{N}_m \right)
\end{equation}
Assuming minimal overlap, the upper bound for receptive field size after $L$ layers is:
\begin{equation}
    |R_n^{(L)}| \leq 1 + w \cdot L
\end{equation}
for a single group. For $g$ groups, the expansion is:
\begin{equation}
    |\mathcal{R}_n^{1D}| \leq \min \left\{ N, g \cdot w \cdot L \right\}
\end{equation}
\begin{equation}
    |\mathcal{R}_n^{2D}| \leq \min \left\{ N, w \cdot L \right\}
\end{equation}}

{\textbf{Coverage ratio:}
\begin{equation}
    \frac{|\mathcal{R}_n^{1D}|}{|\mathcal{R}_n^{2D}|} = g
\end{equation}
This shows that, for the same $w$ and $L$, 1D attention with $g$ groups covers $g$ times more tokens than 2D attention.}

{\textbf{Extreme case (full coverage):}  
If $g \cdot w \cdot L \geq N$, then $|\mathcal{R}_n^{1D}| = N$, i.e., the receptive field covers the entire image in fewer layers.}

{\textbf{Growth curves:}  
For 2D attention, receptive field growth is linear:
\begin{equation}
    f_{2D}(L) = w \cdot L
\end{equation}
For 1D attention, receptive field growth is:
\begin{equation}
    f_{1D}(L) = g \cdot w \cdot L
\end{equation}
Thus,
\begin{equation}
    f_{1D}(L) = g \cdot f_{2D}(L)
\end{equation}}

{\textbf{Graph-theoretic interpretation:}  
The attention connectivity graph for 1D group/dilated attention is denser due to interleaved group assignments, leading to faster expansion of reachable nodes (tokens) compared to the grid-constrained 2D case.}

{\textbf{Spatial distribution:}  
1D attention, through group/dilation, can connect tokens across the entire axis in $O(L)$ steps, while 2D attention requires $O(\sqrt{N}/w)$ steps to reach distant corners on the grid.}

{\textbf{(2) Spatial Structure Restoration via 2D Euclidean Decay}}

{While 1D attention alone may disrupt local spatial continuity, our proposed 2D Euclidean (L2) spatial decay matrix:
\begin{equation}
    E^{2d}_{nm} = \gamma^{\sqrt{(x_n - x_m)^2 + (y_n - y_m)^2}}
\end{equation}
modulates the attention weights as:
\begin{equation}
    A_{nm} = \frac{\exp(Q_n K_m^T/\sqrt{d}) \cdot E^{2d}_{nm}}{\sum_{k \in \mathcal{R}_n^{1D}} \exp(Q_n K_k^T/\sqrt{d}) \cdot E^{2d}_{nk}}
\end{equation}
This spatial prior ensures that, even within a large and irregular receptive field, the model remains sensitive to geometric proximity and local continuity, effectively restoring spatial structure.}

{\textbf{(3) Theoretical Impact of 2D Spatial Prior on 1D Attention}}

{The gradient of attention weights with respect to spatial coordinates is:
\begin{equation}
    \frac{\partial A_{nm}}{\partial x_n} = \log \gamma \cdot A_{nm} \left( \frac{x_n - x_m}{d_{nm}} - \sum_{k \in \mathcal{R}_n^{1D}} A_{nk} \frac{x_n - x_k}{d_{nk}} \right)
\end{equation}
where $d_{nm} = \sqrt{(x_n - x_m)^2 + (y_n - y_m)^2}$. This smooth, direction-aware gradient allows the model to optimize spatial relationships robustly, even when the underlying attention is 1D.}

{Moreover, the spatial entropy of the attention distribution:
\begin{equation}
    H(A_n) = -\sum_{m \in \mathcal{R}_n^{1D}} A_{nm} \log A_{nm}
\end{equation}
is maximized under isotropic constraints, enabling richer spatial representations.}

\begin{table*}[t]
 \centering
 \setlength{\tabcolsep}{2.2mm}
\scalebox{0.88}{
\begin{tabular}{ccccccc}
\toprule
Output Size & Layer Name & EVT-T & EVT-S & EVT-B & EVT-L & EVT-XL \\
\midrule
$56\times 56$ & \begin{tabular}{c} Conv Stem  \end{tabular} & $\begin{array}{c} 3\times3, 32, \text{stride}~2 \\ 3\times3, 32 \\ 3\times3, 32 \\ 3\times3, 64, \text{stride}~2    \end{array}$ & $\begin{array}{c} 3\times3, 32, \text{stride}~2 \\ 3\times3, 32 \\ 3\times3, 32 \\ 3\times3, 64, \text{stride}~2    \end{array}$ & $\begin{array}{c} 3\times3, 40, \text{stride}~2 \\ 3\times3, 40 \\ 3\times3, 40 \\ 3\times3, 80, \text{stride}~2   \end{array}$ & $\begin{array}{c} 3\times3, 56, \text{stride}~2 \\ 3\times3, 56 \\ 3\times3, 56 \\ 3\times3, 112, \text{stride}~2   \end{array}$ & $\begin{array}{c} 3\times3, 64, \text{stride}~2 \\ 3\times3, 64 \\ 3\times3, 64 \\ 3\times3, 128, \text{stride}~2  \end{array}$\\ 
\midrule
\begin{tabular}{c} Stage 1 \end{tabular} & \begin{tabular}{c}CPE \\${\rm EuSA_g/EuSA_d}$\\ FFN\end{tabular} & $\begin{bmatrix}\setlength{\arraycolsep}{1pt} \begin{array}{c}
		3\times3, 64\\ \text{k=}98, \text{h=}2\\ \text{r=}3
		\end{array} \end{bmatrix} \times 2$ & $\begin{bmatrix}\setlength{\arraycolsep}{1pt} \begin{array}{c}
		3\times3, 64\\ \text{k=}98, \text{h=}2\\ \text{r=}3
		\end{array} \end{bmatrix} \times 4$ &  $\begin{bmatrix}\setlength{\arraycolsep}{1pt} \begin{array}{c}
		3\times3, 80\\ \text{k=}98, \text{h=}2\\ \text{r=}3
		\end{array} \end{bmatrix} \times 4$   &  $\begin{bmatrix}\setlength{\arraycolsep}{1pt} \begin{array}{c}
		3\times3, 112\\ \text{k=}98, \text{h=}4\\ \text{r=}3
		\end{array} \end{bmatrix} \times 4$ & $\begin{bmatrix}\setlength{\arraycolsep}{1pt} \begin{array}{c}
		3\times3, 128\\ \text{k=}98, \text{h=}4\\ \text{r=}3
		\end{array} \end{bmatrix} \times 6$ \\
\midrule
$28\times 28$ & Patch Merging & $3\times3, 128, \text{stride}~2$ & $3\times3, 128, \text{stride}~2$ & $3\times3, 160, \text{stride}~2$ &  $3\times3, 224, \text{stride}~2$ & {$3\times3, 256, \text{stride}~2$}\\
\midrule
\begin{tabular}{c} Stage 2\end{tabular} & \begin{tabular}{c}CPE \\${\rm EuSA_g/EuSA_d}$\\ FFN\end{tabular} & $\begin{bmatrix}\setlength{\arraycolsep}{1pt} \begin{array}{c}
		3\times3, 128\\ \text{k=}98, \text{h=}4\\ \text{r=}3
		\end{array} \end{bmatrix} \times 2$ & $\begin{bmatrix}\setlength{\arraycolsep}{1pt} \begin{array}{c}
		3\times3, 128\\ \text{k=}98, \text{h=}4\\ \text{r=}3
		\end{array} \end{bmatrix} \times 4$ & $\begin{bmatrix}\setlength{\arraycolsep}{1pt} \begin{array}{c}
		3\times3, 160\\ \text{k=}98, \text{h=}4\\ \text{r=}3
		\end{array} \end{bmatrix} \times 8$& $\begin{bmatrix}\setlength{\arraycolsep}{1pt} \begin{array}{c}
		3\times3, 224\\ \text{k=}98, \text{h=}8\\ \text{r=}3
		\end{array} \end{bmatrix} \times 8$& $\begin{bmatrix}\setlength{\arraycolsep}{1pt} \begin{array}{c}
		3\times3, 256\\ \text{k=}98, \text{h=}8\\ \text{r=}3
		\end{array} \end{bmatrix} \times 12$\\
\midrule
$14\times 14$ & Patch Merging & $3\times3, 256, \text{stride}~2$ & $3\times3, 256, \text{stride}~2$ & $3\times3, 320, \text{stride}~2$ & $3\times3, 448, \text{stride}~2$ & $3\times3, 512, \text{stride}~2$ \\
\midrule
\begin{tabular}{c} Stage 3 \end{tabular} & \begin{tabular}{c}CPE \\${\rm EuSA_g/EuSA_d}$\\ FFN\end{tabular} & $\begin{bmatrix}\setlength{\arraycolsep}{1pt} \begin{array}{c}
		3\times3, 256\\ \text{k=}98, \text{h=}8\\ \text{r=}3
		\end{array} \end{bmatrix} \times 9$ &$\begin{bmatrix}\setlength{\arraycolsep}{1pt} \begin{array}{c}
		3\times3, 256\\ \text{k=}98, \text{h=}8\\ \text{r=}3
		\end{array} \end{bmatrix} \times 18$ & $\begin{bmatrix}\setlength{\arraycolsep}{1pt} \begin{array}{c}
		3\times3, 320\\ \text{k=}98, \text{h=}8\\ \text{r=}3
		\end{array} \end{bmatrix} \times 26$ &  $\begin{bmatrix}\setlength{\arraycolsep}{1pt} \begin{array}{c}
		3\times3, 448\\ \text{k=}98, \text{h=}14\\ \text{r=}3
		\end{array} \end{bmatrix} \times 26$ & $\begin{bmatrix}\setlength{\arraycolsep}{1pt} \begin{array}{c}
		3\times3, 512\\ \text{k=}98, \text{h=}16\\ \text{r=}3
		\end{array} \end{bmatrix} \times 28$\\
\midrule
$7\times 7$ & Patch Merging & $3\times3, 512, \text{stride}~2$ & $3\times3, 512, \text{stride}~2$ & $3\times3, 512, \text{stride}~2$ & $3\times3, 640, \text{stride}~2$ & $3\times3, 1024, \text{stride}~2$ \\
\midrule
\begin{tabular}{c} Stage 4 \end{tabular} & \begin{tabular}{c}CPE \\${\rm EuSA}$\\ FFN\end{tabular} & $\begin{bmatrix}\setlength{\arraycolsep}{1pt} \begin{array}{c}
		3\times3, 512\\ \text{k=}49, \text{h=}16\\ \text{r=}3
		\end{array} \end{bmatrix} \times 2$ &$\begin{bmatrix}\setlength{\arraycolsep}{1pt} \begin{array}{c}
		3\times3, 512\\ \text{k=}49, \text{h=}16\\ \text{r=}3
		\end{array} \end{bmatrix} \times 4$ &$\begin{bmatrix}\setlength{\arraycolsep}{1pt} \begin{array}{c}
		3\times3, 512\\ \text{k=}49, \text{h=}16\\ \text{r=}3
		\end{array} \end{bmatrix} \times 9$ & $\begin{bmatrix}\setlength{\arraycolsep}{1pt} \begin{array}{c}
		3\times3, 640\\ \text{k=}49, \text{h=}20\\ \text{r=}3
		\end{array} \end{bmatrix} \times 9$ & $\begin{bmatrix}\setlength{\arraycolsep}{1pt} \begin{array}{c}
		3\times3, 1024\\ \text{k=}49, \text{h=}32\\ \text{r=}3
		\end{array} \end{bmatrix} \times 12$ \\
\midrule

$1\times 1$ & Classifier & \multicolumn{5}{c}{Fully Connected Layer, 1000} \\
\midrule
\multicolumn{2}{c}{\# Params} & $15$ M & $27$ M & $57$ M & $101$ M & $205$ M  \\
\midrule
\multicolumn{2}{c}{\# FLOPs} & $2.5$ G & $4.6$ G & $9.8$ G & $18.2$ G & $36.4$ G  \\
\bottomrule
\end{tabular}}
\caption{Architectures for ImageNet classification with resolution $224\times 224$. k is the number of tokens for each group. h is the number of heads. r is the expand ratio of FFN.}
 \label{tab:arch}
\end{table*}

\begin{table}[t]
\begin{center}
\setlength{\tabcolsep}{1mm}
\scalebox{0.95}{
\begin{tabular}{l|cc}
\toprule[1pt]
Settings & pretrain & finetune \\
\midrule
Input resolution &  224$^2$ &  384$^2$ \\
Batch size & 1024 & 512 \\
Optimizer & AdamW & AdamW \\
Learning rate & 1$\times10^{-3}$ & 1$\times10^{-5}$ \\
LR schedule & cosine & cosine \\
Weight decay & 5$\times10^{-2}$ & 1$\times10^{-8}$ \\
Warmup epochs & 5 & 0 \\
Epochs &  300 & 30 \\
\midrule
Horizontal flip & $\checkmark$ & $\checkmark$ \\
Random resize \& Crop  & $\checkmark$ & $\checkmark$ \\
AutoAugment & $\checkmark$ & $\checkmark$ \\
Mixup alpha & 0.8 & 0.8  \\ 
Cutmix alpha & 1.0 & 1.0  \\
Random erasing prob & 0.25 & 0.25  \\
Color jitter  & 0.4 & 0.4  \\
\midrule
Label smoothing & 0.1 & 0.1 \\
Dropout & $\times$ & $\times$ \\
Droppath rate & 0.1/0.15/0.4/0.55/0.7 & 0.1/0.15/0.4/0.55  \\
Repeated augment & $\times$ & $\times$  \\
Grad clipping & $\times$ & $\times$ \\
\bottomrule[1pt]
\end{tabular}}
\caption{Detailed hyper-parameters for training variants of EVT on ImageNet. }
\label{tab:impl}
\end{center}
\end{table}

\begin{table*}[ht]
    \centering
    \setlength{\tabcolsep}{2.5mm}
    \subfloat{
    \scalebox{0.95}{
    \begin{tabular}{c|c|c c|c}
        \toprule[1pt]
        \makecell{Cost} & Model & \makecell{Parmas\\(M)} & \makecell{FLOPs\\(G)} & \makecell{Top1-acc\\(\%)}\\
        \midrule[0.5pt]
        \multirow{15}{*}{\rotatebox{90}{\makecell{tiny model\\$\sim 2.5$G}}} 
        &{MSVmamba-M~\cite{msvmamba}} & {12} & {1.5} & {79.8} \\
        &RegionViT-T~\cite{regionvit} & 14 & 2.4 & 80.4 \\
        &tiny-MOAT-2~\cite{MOAT} & 10 & 2.3 & 81.0 \\
        &VAN-B1~\cite{VAN} & 14 & 2.5 & 81.1 \\
        &Conv2Former-N~\cite{conv2former} & 15 & 2.2 & 81.5 \\
        &{UniRepLKNet-N~\cite{largekernel}} & {18} & {2.8} & {81.6} \\
        &NAT-M~\cite{NAT} & 20 & 2.7 & 81.8 \\
        &FAT-B2~\cite{FAT} & 14 & 2.0 & 81.9 \\
        &QnA-T~\cite{QnA} & 16 & 2.5 & 82.0 \\
        &GC-ViT-XT~\cite{globalvit} & 20 & 2.6 & 82.0 \\
        &SMT-T~\cite{SMT} & 12 & 2.4 & 82.2 \\
        &{TransNeXt-Micro~\cite{transnext}} & {13} & {2.7} & {82.5} \\
        &{OverLoCK-XT~\cite{overlock}} & {16} & {2.6} & {82.7} \\
        &\cellcolor{gray!30}RMT-T~\cite{fan2023rmt} & \cellcolor{gray!30}14 & \cellcolor{gray!30}2.5 & \cellcolor{gray!30}82.4 \\
        &\cellcolor{gray!30}\textbf{EVT-T} & \cellcolor{gray!30}\textbf{15} & \cellcolor{gray!30}\textbf{2.5} & \cellcolor{gray!30}\textbf{83.0} \\
        \midrule[0.5pt]
        \multirow{24}{*}{\rotatebox{90}{\makecell{small model\\$\sim 4.5$G}}} 
        &DeiT-S~\cite{deit} & 22 & 4.6 & 79.9 \\
        &Swin-T~\cite{SwinTransformer} & 29 & 4.5 & 81.3 \\
        &CrossViT-15~\cite{crossvit} & 27 & 5.8 & 81.5 \\
        &ConvNeXt-T~\cite{convnext} & 29 & 4.5 & 82.1 \\
        &Focal-T~\cite{focal} & 29 & 4.9 & 82.2 \\
        &RegionViT-S~\cite{regionvit} & 31 & 5.3 & 82.6 \\
        &{MSVmamba-T~\cite{msvmamba}} & {33} & {4.6} & {82.8} \\
        &{UniRepLKNet-T~\cite{largekernel}} & {31} & {4.9} & {83.2} \\
        &SG-Former-S~\cite{sgformer} & 23 & 4.8 & 83.2 \\
        &Ortho-S~\cite{Ortho} & 24 & 4.5 & 83.4 \\
        &InternImage-T~\cite{internimage} & 30 & 5.0 & 83.5 \\
        & {MILA-T~\cite{MLLA}} & {25} & {4.2} & {83.5} \\
        &MaxViT-T~\cite{maxvit} & 31 & 5.6 & 83.6 \\
        &FAT-B3~\cite{FAT} & 29 & 4.4 & 83.6 \\
        &{VSSD-T~\cite{vssd}} & {24} & {4.5} & {83.7} \\
        &BiFormer-S~\cite{biformer} & 26 & 4.5 & 83.8 \\
        &{TransNeXt-Tiny~\cite{transnext}} & {28} & {5.7} & {84.0} \\ 
        &\cellcolor{gray!30}RMT-S~\cite{fan2023rmt} & \cellcolor{gray!30}27 & \cellcolor{gray!30}4.5 & \cellcolor{gray!30}84.1 \\
        &\cellcolor{gray!30}\textbf{EVT-S} & \cellcolor{gray!30}\textbf{27} & \cellcolor{gray!30}\textbf{4.6} & \cellcolor{gray!30}\textbf{84.4} \\
        \cmidrule(r){2-5}
        &CvT-13$\uparrow384$~\cite{cvt} & 20 & 16.3 & 83.0 \\
        &CoAtNet-0$\uparrow384$~\cite{coatnet} & 20 & 13.4 & 83.9 \\
        &CSwin-T$\uparrow384$~\cite{cswin} & 23 & 14.0 & 84.3 \\
        &iFormer-S$\uparrow384$~\cite{iformer} & 20 & 16.1 & 84.6 \\
        &\cellcolor{gray!30}\textbf{EVT-S$\uparrow384$} & \cellcolor{gray!30}\textbf{27} & \cellcolor{gray!30}\textbf{15.8} & \cellcolor{gray!30}\textbf{85.3} \\
        \bottomrule[1pt]
    \end{tabular}}}
    \subfloat{
    \scalebox{0.95}{
    \begin{tabular}{c|c|c c|c}
        \toprule[1pt]
        \makecell{Cost} & Model & \makecell{Parmas\\(M)} & \makecell{FLOPs\\(G)} & \makecell{Top1-acc\\(\%)}\\
        \midrule[0.5pt]
        \multirow{17}{*}{\rotatebox{90}{\makecell{base model\\$\sim 9.0$G}}}
        &ConvNeXt-S~\cite{convnext} & 50 & 8.7 & 83.1 \\
        &{UniRepLKNet-S~\cite{largekernel}} & {56} & {9.1} & {83.9} \\
        &Quadtree-B-b4~\cite{quadtree} & 64 & 11.5 & 84.0 \\
        &ScaleViT-B~\cite{ScalableViT} & 81 & 8.6 & 84.1 \\
        &CrossFormer++-B~\cite{crossformer++} & 52 & 9.5 & 84.2 \\
        &BiFormer-B~\cite{biformer} & 57 & 9.8 & 84.3 \\
        &{MILA-S~\cite{MLLA}} & {43} & {7.3} & {84.4} \\
        &iFormer-B~\cite{iformer} & 48 & 9.4 & 84.6 \\
        &SE-CoTNetD-152~\cite{cotnet} & 56 & 26.5 & 84.6 \\
        & {TransNeXt-Small~\cite{transnext}} & {50} & {10.3} & {84.7} \\
        &{OverLoCK-S~\cite{overlock}} & {56} & {9.7} & {84.8} \\
        &\cellcolor{gray!30}RMT-B~\cite{fan2023rmt} & \cellcolor{gray!30}54 & \cellcolor{gray!30}9.7 & \cellcolor{gray!30}85.0 \\
        &\cellcolor{gray!30}\textbf{EVT-B} & \cellcolor{gray!30}\textbf{57} & \cellcolor{gray!30}\textbf{9.8} & \cellcolor{gray!30}\textbf{85.3} \\
        \cmidrule(r){2-5}
        &ViTAEv2-48M$\uparrow384$~\cite{vitaev2} & 49 & 41.1 & 84.7 \\
        &MViTv2-B$\uparrow384$~\cite{mvitv2} & 52 & 36.7 & 85.6 \\
        &iFormer-B$\uparrow384$~\cite{iformer} & 48 & 30.5 & 85.7 \\
        &\cellcolor{gray!30}\textbf{EVT-B$\uparrow384$} & \cellcolor{gray!30}\textbf{57} & \cellcolor{gray!30}\textbf{32.8} & \cellcolor{gray!30}\textbf{86.2} \\
        \midrule[0.5pt]
        \multirow{17}{*}{\rotatebox{90}{\makecell{large model\\$\sim 18.0$G}}}
        &DeiT-B~\cite{deit} & 86 & 17.5 & 81.8 \\
        &Swin-B~\cite{SwinTransformer} & 88 & 15.4 & 83.3 \\
        
        &CrossFormer-L~\cite{crossformer} & 92 & 16.1 & 84.0 \\
        &Ortho-L~\cite{Ortho} & 88 & 15.4 & 84.2 \\
        &DaViT-B~\cite{davit} & 88 & 15.5 & 84.6 \\
        &{VSSD-B~\cite{vssd}} & {89} & {16.1} & {84.7} \\
        &{TransNeXt-Base~\cite{transnext}} & {90} & {18.4} & {84.8} \\
        &MaxViT-B~\cite{maxvit} & 120 & 23.4 & 84.9 \\
        &{OverLoCK-B~\cite{overlock}} & {95} & {16.7} & {85.1} \\
        &{MILA-B~\cite{MLLA}} & {96} & {16.2} & {85.3} \\
        &\cellcolor{gray!30}RMT-L & \cellcolor{gray!30}95 & \cellcolor{gray!30}18.2 & \cellcolor{gray!30}85.5 \\
        &\cellcolor{gray!30}\textbf{EVT-L} & \cellcolor{gray!30}\textbf{101} & \cellcolor{gray!30}\textbf{18.2} & \cellcolor{gray!30}\textbf{85.8} \\
        \cmidrule(r){2-5}
        &Swin-B$\uparrow384$~\cite{SwinTransformer} & 88 & 47.0 & 84.2 \\
        &CSwin-B$\uparrow384$~\cite{cswin} & 78 & 47.0 & 85.4 \\
        &CoAtNet-2$\uparrow384$~\cite{coatnet} & 75 & 49.8 & 85.7 \\
        &iFormer-L$\uparrow384$~\cite{iformer} & 87 & 45.3 & 85.8 \\
        &\cellcolor{gray!30}\textbf{EVT-L$\uparrow384$} & \cellcolor{gray!30}\textbf{101} & \cellcolor{gray!30}\textbf{59.1} & \cellcolor{gray!30}\textbf{86.6} \\
        \midrule[0.5pt]
        \multirow{4}{*}{\rotatebox{90}{\makecell{XL model\\$\sim 35.0$G}}}
        &ConvNeXt-L~\cite{convnext} & 198 & 34.4 & 84.3 \\
        &MaxViT-L~\cite{maxvit} & 212 & 43.9 & 85.1 \\
        &GC ViT-L~\cite{globalvit} & 201 & 32.6 & 85.7 \\
        &\cellcolor{gray!30}\textbf{EVT-XL} & \cellcolor{gray!30}\textbf{205} & \cellcolor{gray!30}\textbf{36.4} & \cellcolor{gray!30}\textbf{86.3} \\
        \bottomrule[1pt]
    \end{tabular}}}
    
    \caption{Comparison with the state-of-the-art on ImageNet-1K classification.}
    \vspace{-3mm}
    \label{tab:ImageNet}
\end{table*}

As shown in Tab.~\ref{tab:abgroup}, we conducted experiments on EVT-S and provided a detailed comparison of different grouping methods. Compared to the 2D grouping/shuffling strategy, the 1D grouping/shuffling strategy offers two key advantages:

\textbf{(a) Lower computational complexity and higher efficiency.} The 2D grouping/shuffling strategy operates along both the Height and Width dimensions, resulting in a more complex memory interaction mechanism. In contrast, the 1D grouping/shuffling strategy is applied solely to the token sequence, eliminating redundant 2D computations and improving overall efficiency. As shown in Tab.~\ref{tab:abgroup}, inference with 1D grouping/shuffling is slightly faster than with 2D grouping/shuffling.

\textbf{(b) Longer Modeling Range.} Compared to the 2D grouping/shuffling strategy, the 1D grouping/shuffling strategy enables a longer effective token interaction range within each group/grid, given the same number of tokens per group. Although it loses the explicit 2D spatial priors, it compensates with an extended token receptive field. As shown in Tab.~\ref{tab:abgroup}, without incorporating Euclidean distance priors, the performance of the 1D strategy is slightly inferior to that of the 2D strategy. However, once the distance prior is introduced, the 1D strategy gains sufficient spatial priors, allowing it to achieve superior performance while maintaining lower computational complexity.

\textbf{(c) Improved Resolution Adaptability.} By using 1D grouping/shuffling, the trained model demonstrates better resolution adaptability. We conduct experiments based on EVT-T, performing direct inference and fine-tuning at two resolutions, 384 and 512. The results, as shown in Tab.~\ref{tab:res}, indicate that 1D grouping/shuffling achieves superior performance.

This grouping method reduces the model complexity to linear. Assuming the number of tokens in each group is $k$, where $ k $ is set as a hyperparameter and remains unchanged within the same task (i.e., at a fixed resolution). However, when the task changes (i.e., the input resolution varies), $ k $ can be manually adjusted. The complexity of ${\rm EuSA_g}$ and ${\rm EuSA_d}$ is:
\begin{equation}
    O((hw/k)*k^2d) = O(hwkd)=O(Nkd).
\end{equation}

We employ LCE to enhance the model's local perception capability, which is a depthwise separable convolution applied to $V$:
\begin{equation}
\label{eq:lce}
\begin{aligned}
    & V = W_v X, \\
    & X = {\rm EuSA}(X) + {\rm LCE}(V).
\end{aligned}
\end{equation}

\subsection{Variants of EVT}
Tab.~\ref{tab:arch} shows the variants of EVT. Similar to our previous work RMT~\cite{fan2023rmt}, we have designed four model variants: EVT-T/S/B/L. In addition, to enable a fair comparison with previous work like Swin, we also designed EVT-Swin-T/S/B. For the EVT-Swin series models, we strictly aligned the configurations of EVT and Swin, except that WSA/SWSA is replaced with ${\rm EuSA_g}$/${\rm EuSA_d}$. For the decay coefficient $\gamma$ in the Euclidean decay matrix, we manually set its value to:
\begin{equation}
    \gamma_n = 1-2^{-3-n}
\end{equation}
Here, $n$ is the index of the head in multi-head attention. Setting different decay coefficients for each head endows the model with the capability to perceive multi-scale information.

\begin{table*}[t]
    \centering
    \setlength{\tabcolsep}{0.25mm}
    \subfloat{
    \scalebox{0.95}{
    \begin{tabular}{c|c c|c c c c c c}
        \toprule[1pt]
         \multirow{2}{*}{Backbone} & \multirow{2}{*}{\makecell{Params\\(M)}} & \multirow{2}{*}{\makecell{FLOPs\\(G)}} & \multicolumn{6}{c}{Mask R-CNN $3\times$+MS}\\
          & & & $AP^b$ & $AP^b_{50}$ & $AP^b_{75}$ & $AP^m$ & $AP^m_{50}$ & $AP^m_{75}$\\
          \midrule[0.5pt]
          MPViT-S~\cite{mpvit} & 43 & 268 & 48.4 & 70.5 & 52.6 & 43.9 & 67.6 & 47.5 \\
          {VSSD-T~\cite{vssd}} & {44} & {265} & {48.8} & {70.4} & {53.4} & {43.6} & {67.6} & {46.9}\\
          SMT-S~\cite{SMT} & 40 & 265 & 49.0 & 70.1 & 53.4 & 43.4 & 67.3 & 46.7\\
          CSWin-T~\cite{cswin} & 42 & 279 & 49.0 & 70.7 & 53.7 & 43.6 & 67.9 & 46.6\\
          InternImage-T~\cite{internimage} & 49 & 270 & 49.1 & 70.4 & 54.1 & 43.7 & 67.3 & 47.3 \\
          {MILA-T~\cite{MLLA}} & {44} & {255} & {48.8} & {71.0} & {53.6} & {43.8} & {68.0} & {46.8} \\
          \rowcolor{gray!30}RMT-S & 46 & 262 & 50.7 & 71.9 & 55.6 & 44.9 & 69.1 & 48.4\\
          \rowcolor{gray!30}\textbf{EVT-S} & \textbf{45} & \textbf{262} & \textbf{51.6} & \textbf{72.4} & \textbf{56.6} & \textbf{45.6} & \textbf{69.8} & \textbf{49.1}\\
          \midrule[0.5pt]
          ConvNeXt-S~\cite{convnext} & 70 & 348 & 47.9 & 70.0 & 52.7 & 42.9 & 66.9 & 46.2 \\
          Swin-S~\cite{SwinTransformer} & 69 & 359 & 48.5 & 70.2 & 53.5 & 43.3 & 67.3 & 46.6 \\
          InternImage-S~\cite{internimage} & 69 & 340 & 49.7 & 71.1 & 54.5 & 44.5 & 68.5 & 47.8 \\
          CSWin-S~\cite{cswin} & 54 & 342 & 50.0 & 71.3 & 54.7 & 44.5 & 68.4 & 47.7 \\
          {MILA-S~\cite{MLLA}} & {63} & {319} & {50.5} & {71.8} & {55.2} & {44.9} & {69.1} & {48.2} \\
          \rowcolor{gray!30}RMT-B & 73 & 373 & 52.2 & 72.9 & 57.0 & 46.1 & 70.4 & 49.9  \\
          \rowcolor{gray!30}\textbf{EVT-B} & \textbf{76} & \textbf{371} & \textbf{53.3} & \textbf{73.9} & \textbf{58.4} & \textbf{46.7} & \textbf{71.2} & \textbf{50.6}\\
          \midrule[0.5pt]
          Swin-B~\cite{SwinTransformer} & 107 & 496 & 48.6 & 70.0 & 53.4 & 43.3 & 67.1 & 46.7 \\
          ViT-Adapter-B~\cite{vitadapter} & 120 & 832 & 49.6 & 70.6 & 54.0 & 43.6 & 67.7 & 46.9 \\
          InternImage-B~\cite{internimage} & 115 & 501 & 50.3 & 71.4 & 55.3 & 44.8 & 68.7 & 48.0 \\
          \rowcolor{gray!30}\textbf{EVT-L} & \textbf{119} & \textbf{550} & \textbf{53.6} &  \textbf{74.1} & \textbf{58.7} & \textbf{47.1} & \textbf{71.5} & \textbf{51.0} \\
          \bottomrule
    \end{tabular}}}
    \subfloat{
    \scalebox{0.95}{
    \begin{tabular}{c|c c|c c c c c c}
        \toprule[1pt]
         \multirow{2}{*}{Backbone} & \multirow{2}{*}{\makecell{Params\\(M)}} & \multirow{2}{*}{\makecell{FLOPs\\(G)}} & \multicolumn{6}{c}{Cascade Mask R-CNN $3\times$+MS}\\
          & & & $AP^b$ & $AP^b_{50}$ & $AP^b_{75}$ & $AP^m$ & $AP^m_{50}$ & $AP^m_{75}$\\
          \midrule[0.5pt]
          {UniRepLKNet-T~\cite{largekernel}} & {89} & {749} & {51.8} & {--} & {--} & {44.9} & {--} & {--} \\
          SMT-S~\cite{SMT} & 78 & 744 & 51.9 & 70.5 & 56.3 & 44.7 & 67.8 & 48.6 \\
          UniFormer-S~\cite{uniformer} & 79 & 747 & 52.1 & 71.1 & 56.6 & 45.2 & 68.3 & 48.9 \\
          CSWin-T~\cite{cswin} & 80 & 757 & 52.5 & 71.5 & 57.1 & 45.3 & 68.8 & 48.9 \\
          \rowcolor{gray!30}RMT-S & 83 & 741 & 53.2 & 72.0 & 57.8 & 46.1 & 69.8 & 49.8\\
          \rowcolor{gray!30}\textbf{EVT-S} & \textbf{83} & \textbf{741} & \textbf{53.8} & \textbf{72.4} & \textbf{58.4} & \textbf{46.5} & \textbf{70.2} & \textbf{50.5}\\
          \midrule[0.5pt]
          Swin-S~\cite{SwinTransformer} & 107 & 838 & 51.9 & 70.7 & 56.3 & 45.0 & 68.2 & 48.8 \\
          NAT-S~\cite{NAT} & 108 & 809 & 51.9 & 70.4 & 56.2 & 44.9 & 68.2 & 48.6 \\
          DAT-S~\cite{dat} & 107 & 857 & 52.7 & 71.7 & 57.2 & 45.5 & 69.1 & 49.3 \\
          {UniRepLKNet-S~\cite{largekernel}} & {113} & {835} & {53.0} & {--} & {--} & {45.9} & {--} & {--} \\
          {OverLoCK-S~\cite{overlock}} & {114} & {857} & {53.6} & {--} & {--} & {46.4} & {--} & {--} \\
          UniFormer-B~\cite{uniformer} & 107 & 878 & 53.8 & 72.8 & 58.5 & 46.4 & 69.9 & 50.4 \\
          \rowcolor{gray!30}RMT-B & 111 & 852 & 54.5 & 72.8 & 59.0 & 47.2 & 70.5 & 51.4  \\
          \rowcolor{gray!30}\textbf{EVT-B} & \textbf{114} & \textbf{849} & \textbf{55.5} & \textbf{74.0} & \textbf{60.2} & \textbf{47.9} & \textbf{71.7} & \textbf{52.2}  \\
          \midrule
          Swin-B~\cite{SwinTransformer} & 145 & 982 & 51.9 & 70.5 & 56.4 & 45.0 & 68.1 & 48.9 \\
          GC ViT-B~\cite{globalvit} & 146 & 1018 & 52.9 & 71.7 & 57.8 & 45.8 & 69.2 & 49.8 \\
          {OverLoCK-B~\cite{overlock}} & {154} & {1008} & {53.9} & {--} & {--} & {46.8} & {--} & {--} \\
          CSWin-B~\cite{cswin} & 135 & 1004 & 53.9 & 72.6 & 58.5 & 46.4 & 70.0 & 50.4 \\
          \rowcolor{gray!30}\textbf{EVT-L} & \textbf{157} & \textbf{1029} & \textbf{55.8} & \textbf{74.3} & \textbf{60.4} & \textbf{48.2} & \textbf{72.2} & \textbf{52.2} \\
          \bottomrule
    \end{tabular}}}
    \vspace{-1mm}
    \caption{Comparison with other backbones using "$3\times+\mathrm{MS}$`` schedule on COCO.}
    \label{tab:COCO3x}
    \vspace{-0mm}
\end{table*}
\begin{table*}[h]
    \setlength{\tabcolsep}{0.8mm}
    \centering
    \vspace{-1mm}
    \scalebox{0.95}{
    \begin{tabular}{c|c c|c c c c c c|c c|c c c c c c}
        \toprule[1pt]
        \multirow{2}{*}{Backbone} & \multirow{2}{*}{\makecell{Params\\(M)}} & \multirow{2}{*}{\makecell{FLOPs\\(G)}} & \multicolumn{6}{c|}{Mask R-CNN $1\times$} & \multirow{2}{*}{\makecell{Params\\(M)}} & \multirow{2}{*}{\makecell{FLOPs\\(G)}} & \multicolumn{6}{c}{RetinaNet $1\times$}\\
         & & & $AP^b$ & $AP^b_{50}$ & $AP^b_{75}$ & $AP^m$ & $AP^m_{50}$ & $AP^m_{75}$ & & & $AP^b$ & $AP^b_{50}$ & $AP^b_{75}$ & $AP^b_S$ & $AP^b_{M}$ & $AP^b_{L}$ \\
         \midrule[0.5pt]
        MPViT-XS~\cite{mpvit} & 30 & 231 & 44.2 & 66.7 & 48.4 & 40.4 & 63.4 & 43.4 & 20 & 211 & 43.8 & 65.0 & 47.1 & 28.1 & 47.6 & 56.5 \\
        {VSSD-M~\cite{vssd}} & {33} & {220} & {45.4} & {67.5} & {49.8} & {41.3} & {64.5} & {44.6} & -- & -- & -- & -- & -- & -- & -- & -- \\
        FAT-B2~\cite{FAT} & 33 & 215 & 45.2 & 67.9 & 49.0 & 41.3 & 64.6 & 44.0 & 23 & 196 & 44.0 & 65.2 & 47.2 & 27.5 & 47.7 & 58.8 \\
        \rowcolor{gray!30}RMT-T~\cite{fan2023rmt} & 33 & 218 & 47.1 & 68.8 & 51.7 & 42.6 & 65.8 & 45.9 & 23 & 199 & 45.1 & 66.2 & 48.1 & 28.8 & 48.9 & 61.1 \\
        \rowcolor{gray!30}\textbf{EVT-T} & \textbf{34} & \textbf{221} & \textbf{47.8} & \textbf{69.4} & \textbf{52.2} & \textbf{42.9} & \textbf{66.3} & \textbf{46.4} & \textbf{24} & \textbf{202} & \textbf{46.2} & \textbf{67.3} & \textbf{49.4} & \textbf{29.6} & \textbf{50.7} & \textbf{61.8} \\
        \midrule[0.5pt]
        CMT-S~\cite{cmt} & 45 & 249 & 44.6 & 66.8 & 48.9 & 40.7 & 63.9 & 43.4 & 44 & 231 & 44.3 & 65.5 & 47.5 & 27.1 & 48.3 & 59.1 \\
        CrossFormer-S~\cite{crossformer} & 50 & 301 & 45.4 & 68.0 & 49.7 & 41.4 & 64.8 & 44.6 & 41 & 272 & 44.4 & 65.8 & 47.4 & 28.2 & 48.4 & 59.4 \\
        STViT-S~\cite{stvit} & 44 & 252 & 47.6 & 70.0 & 52.3 & 43.1 & 66.8 & 46.5 & -- & -- & -- & -- & -- & -- & -- & -- \\
        BiFormer-S~\cite{biformer} & -- & -- & 47.8 & 69.8 & 52.3 & 43.2 & 66.8 & 46.5 & -- & -- & 45.9 & 66.9 & 49.4 & 30.2 & 49.6 & 61.7 \\
        {MSVMamba-T~\cite{msvmamba}} & {53} & {252} & {46.9} & {68.8} & {51.4} & {42.2} & {65.6} & {45.4} & -- & -- & -- & -- & -- & -- & -- & -- \\
        {MILA-S~\cite{MLLA}} & {63} & {319} & {49.2} & {71.5} & {53.9} & {44.2} & {68.5} & {47.2} & -- & -- & -- & -- & -- & -- & -- & -- \\
        {TransNeXt-T~\cite{transnext}} & {48} & {269} & {49.9} & {71.5} & {54.9} & {44.6} & {68.6} & {48.1} & -- & -- & -- & -- & -- & -- & -- & -- \\
        \rowcolor{gray!30}RMT-S~\cite{fan2023rmt} & 46 & 262 & 49.0 & 70.8 & 53.9 & 43.9 & 67.8 & 47.4 & 36 & 244 & 47.8 & 69.1 & 51.8 & 32.1 & 51.8 & 63.5 \\
        \rowcolor{gray!30}\textbf{EVT-S} & \textbf{45} & \textbf{262} & \textbf{49.9} & \textbf{71.3} & \textbf{54.9} & \textbf{44.5} & \textbf{68.4} & \textbf{48.1} & \textbf{36} & \textbf{243} & \textbf{48.0} & \textbf{69.2} & \textbf{51.5} & \textbf{30.7} & \textbf{52.5} & \textbf{64.0} \\
        \midrule[0.5pt]
        Swin-S~\cite{SwinTransformer} & 69 & 359 & 45.7 & 67.9 & 50.4 & 41.1 & 64.9 & 44.2 & 60 & 339 & 44.5 & 66.1 & 47.4 & 29.8 & 48.5 & 59.1 \\
        CSWin-S~\cite{cswin} & 54 & 342 & 47.9 & 70.1 & 52.6 & 43.2 & 67.1 & 46.2 & -- & -- & -- & -- & -- & -- & -- & -- \\
        BiFormer-B~\cite{biformer} & -- & -- & 48.6 & 70.5 & 53.8 & 43.7 & 67.6 & 47.1 & -- & -- & 47.1 & 68.5 & 50.4 & 31.3 & 50.8 & 62.6 \\
        {OverLoCK-S~\cite{overlock}} & {75} & {366}  & {49.4} & -- & -- & {44.0} & -- & -- & -- & -- & -- & -- & -- & -- & -- & -- \\
        \rowcolor{gray!30}RMT-B~\cite{fan2023rmt} & 73 & 373 & 51.1 & 72.5 & 56.1 & 45.5 & 69.7 & 49.3 & 63 & 355 & 49.1 & 70.3 & 53.0 & 32.9 & 53.2 & 64.2 \\
        \rowcolor{gray!30}\textbf{EVT-B} & \textbf{76} & \textbf{371} & \textbf{51.7} & \textbf{73.1} & \textbf{56.6} & \textbf{45.8} & \textbf{70.2} & \textbf{49.6} & \textbf{66} & \textbf{352} & \textbf{49.3} & \textbf{70.5} & \textbf{53.1} & \textbf{31.9} & \textbf{53.9} & \textbf{65.5} \\
        \midrule[0.5pt]
        MPViT-B~\cite{mpvit} & 95 & 503 & 48.2 & 70.0 & 52.9 & 43.5 & 67.1 & 46.8 & 85 & 482 & 47.0 & 68.4 & 50.8 & 29.4 & 51.3 & 61.5 \\
        CSwin-B~\cite{cswin} & 97 & 526 & 48.7 & 70.4 & 53.9 & 43.9 & 67.8 & 47.3 & -- & -- & -- & -- & -- & -- & -- & -- \\
        InternImage-B~\cite{internimage} & 115 & 501 & 48.8 & 70.9 & 54.0 & 44.0 & 67.8 & 47.4 & -- & -- & -- & -- & -- & -- & -- & -- \\
        MILA-B~\cite{MLLA} & {115} & {502} & {50.5} & {72.0} & {55.4} & {45.0} & {69.3} & {48.6} & -- & -- & -- & -- & -- & -- & -- & -- \\
        \rowcolor{gray!30}RMT-L & 114 & 557 & 51.6 & 73.1 & 56.5 & 45.9 & 70.3 & 49.8 & 104 & 537 & 49.4 & 70.6 & 53.1 & 34.2 & 53.9 & 65.2 \\
        \rowcolor{gray!30}\textbf{EVT-L} & \textbf{119} & \textbf{550} & \textbf{52.2} & \textbf{73.6} & \textbf{57.5} & \textbf{46.2} & \textbf{70.6} & \textbf{49.9} & \textbf{110} & \textbf{531} & \textbf{50.1} & \textbf{71.4} & \textbf{53.7} & \textbf{33.8} & \textbf{54.9} & \textbf{66.1}\\
        \bottomrule[1pt]
    \end{tabular}}
    \caption{Comparison to other backbones using "$1\times$`` schedule on COCO.}
    \vspace{-3mm}
    \label{tab:COCO1x}
\end{table*}

\section{experiments}

We conducted extensive experiments on the image classfication, object detection, instance segmentation and semantic segmentation. We also evaluate the EVT's robustness on ImageNet-v2/A/R~\cite{imagenetv2, imagenet-a, imagenet-r}. In addition to these experiments, we also conducted detailed ablation studies to verify the role of each module in EVT.

\subsection{Image Classification}
\textbf{Settings: }We use the widely used ImageNet-1K~\cite{imagenet} to conduct the image classification task. The dataset contains 1.28M images for training and 50K images for validation. The detailed settings for pretraining and finetuning are listed in the Tab.~\ref{tab:impl}. The AdamW is used with a cosine decay learning rate scheduler. The initial learning rate,  weight decay, and  batch-size are set to  0.001, 0.05, and 1024, respectively. We apply the same data augmentation and regularization used  in  DeiT~\cite{deit} (RandAugment \cite{randomaugment} (randm9-mstd0.5-inc1) , Mixup \cite{mixup} (prob = 0.8), CutMix \cite{cutmix} (prob = 1.0), Random Erasing (prob = 0.25), and Exponential Moving Average (EMA) \cite{EMA}). The  maximum rates of increasing stochastic depth \cite{droppath} are set to 0.1/0.15/0.4/0.55 for EVT-T/S/B/L.

\noindent\textbf{Results. }We compare our EVT against the state-of-the-art models in Tab.~\ref{tab:ImageNet}.  The comparison results demonstrate that our EVT outperforms previous models under different settings in terms of FLOPs and model size. Specifically, our EVT-T achieves \textbf{83.0\%} top-1 acc, surpasses RMT-T by \textbf{0.6\%} with similar Params and FLOPs. It achieved the same accuracy as MPViT-S, but requires only about half the computation. This fully demonstrates the advantages of EVT. Our EVT-B achieves \textbf{85.3\%} top1-acc, not only surpassing the corresponding counterparts with the base model size but also surpassing those with large model size. As for the resolution of $384\times 384$, our large model EVT-L achieves an accuracy of \textbf{86.6\%}. This result significantly surpasses its counterparts.

\begin{table*}[t]
    \centering
    \setlength{\tabcolsep}{2.3mm}
    \subfloat{
    \scalebox{0.95}{
    \begin{tabular}{c|c c|c}
         \toprule[1pt]
         \multicolumn{4}{c}{Semantic FPN}\\
         \midrule[0.5pt]
         Backbone & Params(M) & FLOPs(G) & mIoU(\%)\\
         \midrule[0.5pt]
         PVTv2-B1~\cite{pvtv2}& 18 & 34 & 42.5 \\
         FAT-B2~\cite{FAT} & 17 & 32 & 45.4\\
         EdgeViT-S~\cite{edgevit}& 17 & 32 & 45.9\\
         \rowcolor{gray!30}RMT-T~\cite{fan2023rmt}& 17 & 34 & 46.4\\
         \rowcolor{gray!30}\textbf{EVT-T}& \textbf{18} & \textbf{34} & \textbf{48.3}\\
         \midrule[0.5pt]
         DAT-T~\cite{dat}& 32 & 198 & 42.6 \\
         RegionViT-S+~\cite{regionvit} & 35 & 236 & 45.3 \\
         CSWin-T~\cite{cswin}& 26 & 202 & 48.2 \\
         FAT-B3~\cite{FAT} & 33 & 179 & 48.9 \\
         \rowcolor{gray!30}RMT-S~\cite{fan2023rmt}& 30 & 180 & 49.4\\
         \rowcolor{gray!30}\textbf{EVT-S}& \textbf{30} & \textbf{180} & \textbf{50.0}\\
         \midrule[0.5pt]
         DAT-S~\cite{dat}& 53 & 320 & 46.1\\
         RegionViT-B+~\cite{regionvit} & 77 & 459 & 47.5 \\
         UniFormer-B~\cite{uniformer}& 54 & 350 & 47.7 \\
         CSWin-S~\cite{cswin}& 39 & 271 & 49.2 \\
         \rowcolor{gray!30}RMT-B~\cite{fan2023rmt}& 57 & 294 & 50.4 \\
         \rowcolor{gray!30}\textbf{EVT-B}& \textbf{60} & \textbf{291} & \textbf{51.7} \\
         \midrule[0.5pt]
         DAT-B~\cite{dat}& 92 & 481 & 47.0 \\
         CrossFormer-L~\cite{crossformer}& 95 & 497 & 48.7 \\
         PVTv2-B5~\cite{pvtv2} & 86 & 486 & 48.7 \\
         CSWin-B~\cite{cswin}& 81 & 464 & 49.9 \\
         \rowcolor{gray!30}RMT-L~\cite{fan2023rmt}& 98 & 482 & 51.4 \\
         \rowcolor{gray!30}\textbf{EVT-L}& \textbf{103} & \textbf{475} & \textbf{52.0} \\
         \bottomrule[1pt]
    \end{tabular}}}
    \subfloat{
    \scalebox{0.95}{
    \begin{tabular}{c|c c|c}
         \toprule[1pt]
         \multicolumn{4}{c}{UperNet}\\
         \midrule[0.5pt]
         Backbone & Params(M) & FLOPs(G) & mIoU(\%)\\
         \midrule[0.5pt]
         NAT-T~\cite{NAT}& 58 & 934 & 47.1 \\
         {MSVmamba-T~\cite{msvmamba}} & {65} & {942} & {47.6} \\
         {VSSD-T~\cite{vssd}} & {53} & {941} & {47.9} \\
         MPViT-S~\cite{mpvit}& 52 & 943 & 48.3 \\
         {UniRepLKNet-T~\cite{largekernel}} & {61} & {946} & {48.6} \\
         HorNet-T~\cite{hornet}& 55 & 924 & 49.2 \\
         SMT-S~\cite{SMT}& 50 & 935 & 49.2 \\
         \rowcolor{gray!30}RMT-S~\cite{fan2023rmt}& 56 & 937 & 49.8 \\
         \rowcolor{gray!30}\textbf{EVT-S}& \textbf{56} & \textbf{936} & \textbf{49.8} \\
         \midrule[0.5pt]
         NAT-S~\cite{NAT} & 82 & 1010 & 48.0 \\
         DAT-S~\cite{dat}& 81 & 1079 & 48.3\\
         InterImage-S~\cite{internimage}& 80 & 1017 & 50.2\\
         CSWin-S~\cite{cswin}& 65 & 1027 & 50.4\\
         {OverLoCK-S~\cite{overlock}} & {85} & {1051} & {51.3} \\
         {TransNext-S~\cite{transnext}} & {80} & {1028} & {52.2} \\
         \rowcolor{gray!30}RMT-B~\cite{fan2023rmt}& 83 & 1051 & 52.0\\
         \rowcolor{gray!30}\textbf{EVT-B} & \textbf{86} & \textbf{1048} & \textbf{52.7}\\
         \midrule
         Focal-B~\cite{focal} & 126 & 1354 & 49.0 \\
         DAT-B~\cite{dat} & 121 & 1212 & 49.4 \\
         HorNet-B~\cite{hornet} & 126 & 1171 & 50.5 \\
         {MILA-B~\cite{MLLA}} & {128} & {1183} & {51.9} \\
         {TransNext-B~\cite{transnext}} & {121} & {1265} & {53.0} \\
         \rowcolor{gray!30}RMT-L~\cite{fan2023rmt} & 125 & 1241 & 52.8\\
         \rowcolor{gray!30}\textbf{EVT-L}& \textbf{131} & \textbf{1234} & \textbf{53.6}\\
         \bottomrule[1pt]
    \end{tabular}}}
    \caption{Comparison with the state-of-the-art on ADE20K. }
    \vspace{-3mm}
    \label{tab:ade20k}
\end{table*}

\begin{table*}[ht]
    \centering
    \begin{tabular}{c|cc|cccccc}
    \toprule[1pt]
         Backbone & Params(M) & FLOPs(G) & IN-1K$\uparrow$ & IN-C$\downarrow$ & IN-A$\uparrow$ & IN-R$\uparrow$ & Sketch$\uparrow$ & IN-V2$\uparrow$  \\
         \midrule[0.5pt]
         PVTv2-B1~\cite{pvtv2} & 14 & 2.1 & 78.7 & 62.6 & 14.7 & 41.8 & 28.9 & 66.9 \\
         BiFormer-T~\cite{biformer} & 13 & 2.2 & 81.4 & 55.7 & 25.7 & 45.4 & 31.5 & 70.6 \\
         TransNeXt-Micro~\cite{transnext} & 13 & 2.7 & 82.5 & 50.8 & 29.9 & 45.8 & 33.0 & 72.6 \\
         EVT-T & 15 & 2.5 & \textbf{83.0} & \textbf{49.2} & \textbf{34.7} & \textbf{46.7} & \textbf{35.4} & \textbf{72.6} \\
         \midrule[0.5pt]
         Swin-T~\cite{SwinTransformer} & 29 & 4.5 & 81.3 & 62.0 & 21.7 & 41.3 & 29.0 & 69.7 \\
         FocalNet-T~\cite{focalnet} & 29 & 4.5 & 82.3 & 55.0 & 23.5 & 45.1 & 31.8 & 71.2 \\
         BiFormer-S~\cite{biformer} & 26 & 4.5 & 83.8 & 48.5 & 39.5 & 49.6 & 36.4 & 73.7 \\
         TransNeXt-Tiny~\cite{transnext} & 28 & 5.7 & 84.0 & 46.5 & 39.9 & \textbf{49.6} & 37.6 & 73.8 \\
         EVT-S & 27 & 4.6 & \textbf{84.4} & \textbf{44.2} & \textbf{44.2} & 46.7 & \textbf{38.8} & \textbf{74.6} \\
         \midrule[0.5pt]
         BiFormer-B~\cite{biformer} & 57 & 9.8 & 84.3 & 47.2 & 44.3 & 49.7 & 35.3 & 74.0 \\
         MaxViT-Small~\cite{maxvit} & 69 & 11.7 & 84.4 & 46.4 & 40.0 & 50.6 & 38.3 & 74.0 \\
         TransNeXt-Small~\cite{transnext} & 50 & 10.3 & 84.7 & 43.9 & 47.1 & 52.5 & 39.7 & 74.8 \\
         EVT-B & 57 & 9.8 & \textbf{85.3} & \textbf{43.1} & \textbf{51.4} & \textbf{55.2} & \textbf{41.4} & \textbf{75.7}\\
         \midrule[0.5pt]
         Swin-B~\cite{SwinTransformer} & 88 & 15.4 & 83.5 & 54.5 & 35.9 & 46.6 & 32.4 & 72.3 \\
         ConvNeXt-B~\cite{convnext} & 89 & 15.4 & 83.8 & 46.8 & 36.7 & 51.3 & 38.2 & 73.7 \\
         TransNeXt-Base~\cite{transnext} & 90 & 18.4 & 84.8 & 43.5 & 50.6 & 53.9 & 41.4 & 75.1 \\
         EVT-L & 101 & 18.2 & \textbf{85.8} & \textbf{42.7} & \textbf{55.5} & \textbf{56.9} & \textbf{43.2} & \textbf{76.1}\\
         \midrule[0.5pt]
         GC ViT-L~\cite{globalvit} & 201 & 32.6 & 85.7 & 42.9 & 54.8 & 56.0 & 43.4 & 75.5 \\
         EVT-XL & \textbf{205} & \textbf{36.4} & \textbf{86.3} & \textbf{41.1} & \textbf{56.7} & \textbf{58.0} & \textbf{44.0} & \textbf{76.4} \\
    \bottomrule[1pt]
    \end{tabular}
    \caption{Evaluation of the model's robustness.}
    \vspace{-3mm}
    \label{tab:robustness}
\end{table*}

\subsection{Object Detection and Instance Segmentation}

\textbf{Settings: }We evaluate the proposed EVT on the MS-COCO~\cite{coco}, a widely used dataset. It contains 118K training images and 5K validation images. We utilize MMDetection~\cite{mmdetection} to implement Mask-RCNN~\cite{maskrcnn}, Cascade Mask R-CNN~\cite{cai18cascadercnn}, and RetinaNet~\cite{retinanet} for evaluating the proposed EVT. For Mask R-CNN and Cascade Mask R-CNN, we use the commonly used “$3\times + MS$” setting, and for Mask R-CNN and RetinaNet, we apply the “$1\times$” setting. Following previous works~\cite{SwinTransformer, cswin}, during training, we resize the images such that the shorter side is 800 pixels while keeping the longer side within 1333 pixels. We employ the AdamW optimizer for model optimization.

\noindent\textbf{Results. }Tab.~\ref{tab:COCO3x} and Tab.~\ref{tab:COCO1x} shows the results of object detection and instance segmentation. For "$3\times + MS$" schedule, EVT demonstrates significant advantages over its competitors across various model scales. Specifically, EVT-B, compared to RMT-B, achieves an improvement of \textbf{1.1} in $AP^b$ and \textbf{0.6} in $AP^m$ under the Mask R-CNN framework. In the Cascade Mask R-CNN framework, EVT-L outperforms CSwin-B by \textbf{1.8} in both $AP^b$ and $AP^m$.

For "$1\times$" schedule, EVT also demonstrates its advantages. For example, under the Mask R-CNN framework, EVT-S achieves an improvement of \textbf{0.9} in $AP^b$ and \textbf{0.6} in $AP^m$ compared to RMT-S. Additionally, the performance of EVT-B surpasses that of most models an order of magnitude larger. Under the RetinaNet framework, EVT-T achieves an improvement of \textbf{1.1} in $AP^b$ compared to RMT-T. Its performance even surpasses that of the significantly larger STViT-S.

\subsection{Semantic Segmentation}
\noindent\textbf{Settings. }ADE20K~\cite{ade20k} is a popular dataset for semantic
segmentation, which has 20K training examples and 2K validation images. We utilize Semantic FPN~\cite{semanticfpn} and UperNet~\cite{upernet} to evaluate EVT's performance on the semantic segmentation task. We implement the two frameworks based on the MMSegmentation~\cite{mmsegmentation}. The two frameworks both use an encoder-decoder structure where we employ variants of EVT as the encoder in the frameworks. We follow the traing recipes in Swin Transformer~\cite{SwinTransformer} and PVT~\cite{pvt}, initializing the encoder from ImageNet pretrained models. After that, we train 160K iterations for UperNet and 80K for SemanticFPN. All models use an input resolution of $512\times 512$, and during testing, the image's shorter side is resized to 512 pixels.

\noindent\textbf{Results. }Tab.~\ref{tab:ade20k} shows the results of different variants of EVT with two segmentation frameworks on ADE20K. For Semantic FPN, EVT-T/S/B/L achieves the mIoU of 48.3, 50.0, 51.7 and 52.0, respectively, which leads to an impressive performance gain over its counterparts. For UperNet, EVT also demonstrate superiority. Specifically, EVT-B achieves the mIoU of \textbf{52.7}, surpassing previous SOTA RMT-B by \textbf{0.7}. EVT-B's performance even exceeds that of many models which are an order of magnitude larger. In summary, EVT exhibits significant performance advantages across various scales.

{\subsection{Robustness Evaluation}}

\noindent\textbf{Settings. }We evaluate the model's robustness on several popular datasets~\cite{imagenet-c,imagenet-a,imagenet-r,imagenet-sketch}. We also evaluate the overfitting level of EVT on ImageNet-V2~\cite{imagenet-v2}. The models used for evaluation are pretrained on the ImageNet-1K.

{\noindent\textbf{Results. } The robustness evaluation results are shown in Tab.~\ref{tab:robustness}. On ImageNet-V2 (IN-V2), EVT performs better than all its counterparts. For example, EVT-B surpasses BiFormer-B for \textbf{+1.7} with similar parameters and FLOPs. In ImageNet-A (IN-A) and ImageNet-R (IN-R), the advantages demonstrated by EVT become even more pronounced. Specifically, pretrained solely on ImageNet-1k, EVT-L achieves an accuracy of \textbf{55.5} on ImageNet-A and \textbf{56.9} on ImageNet-R.}

\subsection{Efficiency Comparison}

\noindent\textbf{Settings. }We compare the throughput of different models on the same device. Specifically, we test the speed of various models on a single NVIDIA A100 using a batch size of 64 and fp32 precision.

\begin{table}[ht]
    \centering
    \setlength{\tabcolsep}{1.5mm}
    \scalebox{0.95}{
    \begin{tabular}{c|c c c| c}
    \toprule[1pt]
    Model & \makecell{Params\\(M)} & \makecell{FLOPs\\(G)} & \makecell{Top1-acc\\(\%)} & \makecell{Throughput\\(imgs/s)} \\
    \midrule[0.5pt]
    MPViT-XS~\cite{mpvit} & 11 & 2.9 & 80.9 & 1496 \\
    BiFormer-T~\cite{biformer} & 13 & 2.2 & 81.4 & 1602 \\
    SMT-T~\cite{SMT} & 12 & 2.4 & 82.2 & 636 \\
    {VSSD-M~\cite{vssd}} & {14} & {2.3} & {82.5} & {712} \\
    {OverLoCK-XT} & {16} & {2.6} & {82.7} & {678} \\
    \rowcolor{gray!30}RMT-T~\cite{fan2023rmt} & 14 & 2.5 & 82.4 & 1650 \\
    \rowcolor{gray!30}\textbf{EVT-T} & \textbf{15} & \textbf{2.5} & \textbf{83.0} & \textbf{2142} \\
    \midrule[0.5pt]
    CSWin-T~\cite{cswin} & 22 & 4.3 & 82.7 & 1561 \\
    {MSVMamba-T~\cite{msvmamba}} & {33} & {4.6} & {82.8} & {436}\\
    {MILA-T~\cite{MLLA}} & {25} & {4.2} & {83.5} & {1203} \\
    MaxViT-T~\cite{maxvit} & 31 & 5.6 & 83.6 & 826 \\
    SMT-S~\cite{SMT} & 20 & 4.8 & 83.7 & 356 \\
    BiFormer-S~\cite{biformer} & 26 & 4.5 & 83.8 & 766 \\
    {OverLoCK-T~\cite{overlock}} & {33} & {5.5} & {84.2} & {382} \\
    \rowcolor{gray!30}RMT-Swin-T~\cite{fan2023rmt} & 29 & 4.7 & 83.6 & 1192 \\
    \rowcolor{gray!30}\textbf{EVT-Swin-T} & \textbf{29} & \textbf{4.8} & \textbf{83.9} & \textbf{1420} \\
    \rowcolor{gray!30}RMT-S~\cite{fan2023rmt} & 27 & 4.5 & 84.1 & 876 \\
    \rowcolor{gray!30}\textbf{EVT-S} & \textbf{27} & \textbf{4.6} & \textbf{84.4} & \textbf{1001} \\
    \midrule[0.5pt]
    Swin-S~\cite{SwinTransformer} & 50 & 8.8 & 83.0 & 1006 \\
    {UniRepLKNet-S~\cite{largekernel}} & {56} & {9.1} & {83.9} & {998}\\
    BiFormer-B~\cite{biformer} & 57 & 9.8 & 84.3 & 518 \\
    MaxViT-S~\cite{maxvit} & 69 & 11.7 & 84.5 & 546 \\
    iFormer-B~\cite{iformer} & 48 & 9.4 & 84.6 & 688 \\
    {TransNeXt-Small~\cite{transnext}} & {50} & {10.3} & {84.7} & {302}\\
    {OverLoCK-S~\cite{overlock}} & {56} & {9.7} & {84.8} & {241} \\
    \rowcolor{gray!30}RMT-Swin-S~\cite{fan2023rmt} & 50 & 9.1 & 84.5 & 722 \\
    \rowcolor{gray!30}\textbf{EVT-Swin-S} & \textbf{50} & \textbf{9.3} & \textbf{85.1} & \textbf{866} \\
    \rowcolor{gray!30}RMT-B~\cite{fan2023rmt} & 54 & 9.7 & 85.0 & 457 \\
    \rowcolor{gray!30}\textbf{EVT-B} & \textbf{57} & \textbf{9.8} & \textbf{85.3} & \textbf{502} \\
    \midrule[0.5pt]
    CSWin-B~\cite{cswin} & 78 & 15.0 & 84.2 & 660 \\
    MPViT-B~\cite{mpvit} & 75 & 16.4 & 84.3 & 498 \\
    SMT-L~\cite{SMT} & 80 & 17.7 & 84.6 & 158 \\
    {TransNeXt-Base~\cite{transnext}} & {90} & {18.4} & {84.8} & {202} \\
    {MILA-B~\cite{MLLA}} & {96} & {16.2} & {85.3} & {466} \\
    {VSSD-B~\cite{vssd}} & {89} & {16.1} & {85.4} & {289} \\
    \rowcolor{gray!30}\textbf{EVT-Swin-B} & \textbf{88} & \textbf{16.1} & \textbf{85.3} & \textbf{700} \\
    \rowcolor{gray!30}RMT-L~\cite{fan2023rmt} & 95 & 18.2 & 85.5 & 326 \\
    \rowcolor{gray!30}\textbf{EVT-L} & \textbf{101} & \textbf{18.2} & \textbf{85.8} & \textbf{368} \\
    \midrule[0.5pt]
    MaxViT-L~\cite{maxvit} & 212 & 43.9 & 85.1 & 127 \\
    GC ViT-L~\cite{globalvit} & 201 & 32.6 & 85.7 & 138 \\
    \rowcolor{gray!30}\textbf{EVT-XL} & \textbf{205} & \textbf{36.4} & \textbf{86.3} & \textbf{142} \\
    \bottomrule[1pt]
         
    \end{tabular}}
    \caption{Inference throughputs of different models. Models are tested on resolution of $224\times224$. We benchmark the throughputs on a single NVIDIA A100 GPU with batch size of 64. }
    \label{tab:inference}
\end{table}

\noindent\textbf{Results. }We present the efficiency comparison of different models in Tab.~\ref{tab:inference}, where EVT demonstrates the best accuracy-speed trade-off. Specifically, compared to its baseline model RMT, EVT achieves efficiency improvements at all scales, along with better classification accuracy. Compared to other state-of-the-art models such as BiFormer, EVT not only has faster inference speed (1001 v.s. 766), but also achieves a significant improvement in classification accuracy (84.4 v.s. 83.8). These results fully illustrate the advantages of EVT in both performance and speed.

\subsection{Ablation Study}

\begin{table*}[ht]
    \centering
    \setlength{\tabcolsep}{2mm}
    \scalebox{1.0}{
    \begin{tabular}{c| c |c c|c|c c|c}
        \toprule[1pt]
         Model & Throughput(imgs/s) & Params(M) & FLOPs(G) & Top1-acc(\%) & $AP^b$ & $AP^m$ & mIoU(\%)\\
         \midrule[0.5pt]
         Swin-T~\cite{SwinTransformer}& 1704 & 29 & 4.5 & 81.3 & 43.7 & 39.8 & 44.5 \\
         RMT-Swin-T & 1192 & 29 & 4.7 & 83.6(\textcolor{red}{+2.3}) & 47.8(\textcolor{red}{+4.1}) & 43.1(\textcolor{red}{+3.3}) & 49.1(\textcolor{red}{+4.6}) \\
         EVT-Swin-T & 1420 & 29 & 4.8 & 83.9(\textcolor{red}{+2.6}) & 48.6(\textcolor{red}{+4.9}) & 43.9(\textcolor{red}{+4.1}) & 49.6(\textcolor{red}{+5.1}) \\
         \midrule[0.5pt]
         Swin-S~\cite{SwinTransformer} & 1006 & 50 & 8.8 & 83.0 & 45.7 & 41.1 & 47.6 \\
         RMT-Swin-S & 722 & 50 & 9.1 & 84.5(\textcolor{red}{+1.5})& 49.5(\textcolor{red}{+3.8}) & 44.2(\textcolor{red}{+3.1}) & 51.0(\textcolor{red}{+3.4})\\
         EVT-Swin-S & 866 & 50 & 9.3 & 85.1(\textcolor{red}{+2.1}) & 50.5(\textcolor{red}{+4.8}) & 45.1(\textcolor{red}{+4.0}) & 51.7(\textcolor{red}{+4.1}) \\
         \midrule[0.5pt]
         Swin-B~\cite{SwinTransformer}& 798 & 88 & 15.5 & 83.5 & -- & -- & -- \\
         EVT-Swin-B & 700 & 88 & 16.1 & 85.3(\textcolor{red}{+1.8}) & -- & -- & -- \\
         \midrule[0.5pt]
         EVT-T & 2142 & 15 & 2.5 & 83.0 & 47.8 & 42.9 & 48.3 \\
         w/o LCE & 2315 & 15 & 2.5 & 82.8(\textcolor{brown}{-0.2}) & 47.1(\textcolor{brown}{-0.7}) & 42.3(\textcolor{brown}{-0.6}) & 47.6(\textcolor{brown}{-0.7}) \\
         w/o CPE & 2204 & 15 & 2.5 & 82.9(\textcolor{brown}{-0.1}) & 47.5(\textcolor{brown}{-0.3}) & 42.7(\textcolor{brown}{-0.2}) & 48.0(\textcolor{brown}{-0.3}) \\
         w/o Conv Stem & 2296 & 14 & 2.2 & 82.8(\textcolor{brown}{-0.2}) & 47.3(\textcolor{brown}{-0.5}) & 42.2(\textcolor{brown}{-0.7}) & 47.1(\textcolor{brown}{-1.2}) \\
         ${\rm EuSA_g\xrightarrow{}EuSA_d}$ & 2145 & 15 & 2.5 & 82.5(\textcolor{brown}{-0.5}) & 46.9(\textcolor{brown}{-0.9}) & 41.6(\textcolor{brown}{-1.3}) & 46.8(\textcolor{brown}{-1.5}) \\
         ${\rm EuSA_d\xrightarrow{}EuSA_g}$ & 2145 & 15 & 2.5 & 82.6(\textcolor{brown}{-0.4}) & 47.2(\textcolor{brown}{-0.6}) & 42.0(\textcolor{brown}{-0.9}) & 47.0(\textcolor{brown}{-1.3}) \\
         w/o decay matrix & 2168 & 15 & 2.5 & 82.3(\textcolor{brown}{-0.7}) & 47.1(\textcolor{brown}{-0.7}) & 42.2(\textcolor{brown}{-0.7}) & 46.6(\textcolor{brown}{-1.7}) \\
         \bottomrule[1pt]
    \end{tabular}}
    \vspace{-3mm}
    \caption{Ablation Studies. }
    \vspace{-3mm}
    \label{tab:baseline}
\end{table*}

\begin{figure}[h]
    \centering
    \includegraphics[width=0.85\linewidth]{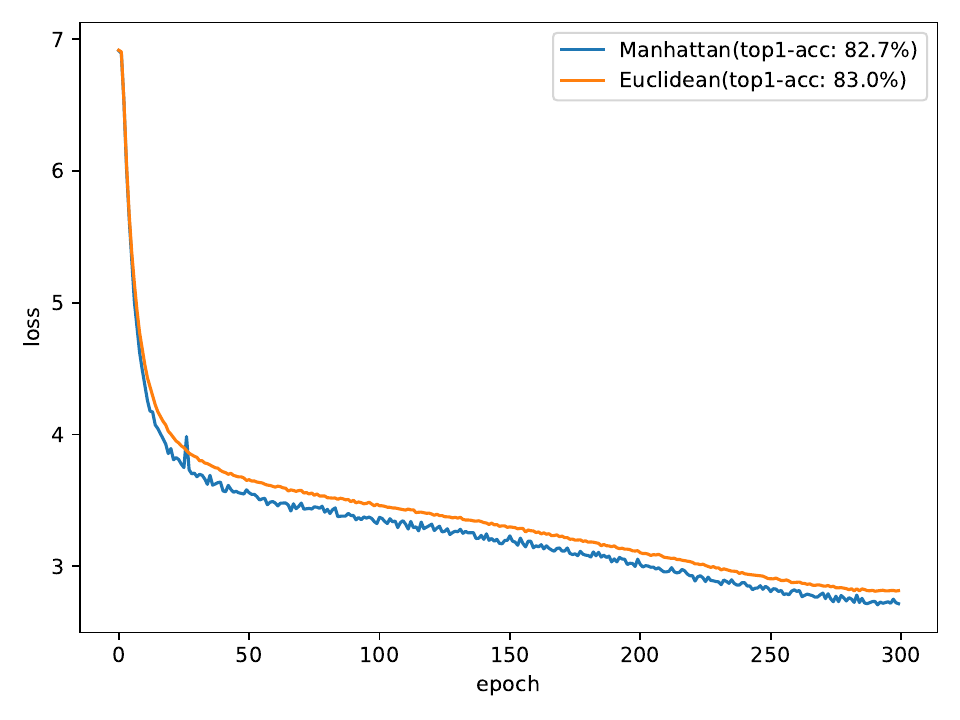}
    \caption{Loss curves for Euclidean distance and Manhattan distance. The experiments are conducted based on EVT-T.}
    \label{fig:loss_curve}
\end{figure}

\noindent\textbf{Comparison between Manhattan distance and Euclidean distance.} Our experiments demonstrate the advantages of Euclidean distance over Manhattan distance. The experiments are based on the EVT-T. As shown in Fig.~\ref{fig:loss_curve}, in addition to achieving better model performance, using Euclidean distance results in reduced overfitting and more stable training.

\noindent\textbf{Comparison among different decay matrix.} Since the gradients of Minkowski distance and RBF are also continuous and preserve directional optimization information, they are theoretically feasible choices for serving as spatial priors. However, in practice, this is not the case. Since our explicit decay is directly applied to the attention scores, it effectively scales them. However, the scaling factor should not be too complex, as excessive scaling can significantly compromise the stability of model training. When using Minkowski distance or radial basis function (RBF) kernels, the high-order power or exponential growth of the distance function can lead to severe instability, which becomes particularly evident in high-resolution tasks such as object detection, where the number of tokens is considerably large. During training, we consistently observed \textbf{distinct loss spikes} and \textbf{Nan} when using Minkowski distance or RBF.  In contrast, no such phenomenon occurred when employing Euclidean distance. As shown in Tab.~\ref{tab:ab_distance}, using specialized distance functions can introduce instability or lead to significant performance degradation.
\begin{table}[h]
    \centering
    \setlength{\tabcolsep}{0.3mm}
    \begin{tabular}{c|c c|c c c}
    \toprule[1pt]
         Model & Params(M) & FLOPs(G) & Acc(\%) & $AP^b$ & $AP^m$ \\
         \midrule
         Euclidean & 29 & 4.8 & 83.9 & 48.6 & 43.9 \\
         \midrule
         Minkowski p=3 & 29 & 4.8 & 83.5 & \makecell{44.4\\(\textcolor{red}{-4.2})} & \makecell{40.6\\(\textcolor{red}{-3.3})} \\ 
         Minkowski p=4 & 29 & 4.8 & Nan & -- & -- \\ 
         Minkowski p=5 & 29 & 4.8 & Nan & -- & -- \\ 
         \midrule
         Gaussian RBF & 29 & 4.8 & Nan & -- & -- \\
         Multiquadric RBF & 29 & 4.8 & 83.6 & \makecell{47.0\\(\textcolor{red}{-1.6})} & \makecell{42.2\\(\textcolor{red}{-1.7})} \\
         Inverse Multiquadric RBF & 29 & 4.8 & Nan & -- & -- \\
         \bottomrule[1pt]
    \end{tabular}
    \caption{Ablation study about the different distance functions. Overly complex distance functions can degrade training stability and lead to performance deterioration.}
    \label{tab:ab_distance}
\end{table}

\noindent\textbf{Strict Comparison with Baselines. }To enable a fair comparison with previous methods, we design the EVT-Swin series models. EVT-Swin strictly aligns with the various configurations of Swin-Transformer, replacing only the WSA/SWSA in Swin-Transformer with our ${\rm EuSA_g}/{\rm EuSA_d}$, and replacing the vanilla attention in Swin-Transformer with our EuSA, while not using additional modules such as CPE and Conv Stem. 

\begin{table}[t]
    \centering
    \setlength{\tabcolsep}{2mm}
    \scalebox{1.}{
    \begin{tabular}{c|c c}
    \toprule[1pt]
         $\gamma_n$ & Acc(\%) & mIoU(\%) \\
         \midrule[0.5pt]
         None & 82.3 & 46.6 \\
         $1-2^{-3}$ & 82.4 & 46.7 \\
         $1-2^{-7}$ & 82.4 & 46.6 \\
         $1-2^{-3-n}$ & 83.0 & 48.3 \\
         $1-2^{-7-n}$ & 82.9 & 47.8 \\
    \bottomrule[1pt]     
    \end{tabular}}
    \caption{Ablation study for $\gamma$.}
    \vspace{-3mm}
    \label{tab:gamma}
\end{table}

\begin{table}[t]
    \centering
    \setlength{\tabcolsep}{1mm}
    \begin{tabular}{c|c c|c c}
    \toprule[1pt]
         Model & \makecell{Params\\(M)} & \makecell{FLOPs\\(G)} & \makecell{Top1-acc\\(\%)} & \makecell{mIoU\\(\%)} \\
         \midrule
         RMT-T & 14 & 2.5 & 82.4 & 46.4 \\
         deeper & 15 & 2.7 & 82.5(\textcolor{red}{+0.1}) & 46.7(\textcolor{red}{+0.3}) \\
         ${\rm MaSA \xrightarrow{}EuSA}$ & 15 & 2.7 & 82.8(\textcolor{red}{+0.4}) & 47.6(\textcolor{red}{+1.2}) \\
         decop$\xrightarrow{}$group & 15 & 2.5 & 83.0(\textcolor{red}{+0.6}) & 48.3(\textcolor{red}{+1.9})\\
    \bottomrule
    \end{tabular}
    \caption{Roadmap from RMT to EVT.}
    \label{tab:roadmap}
\end{table}

\begin{table}[t]
    \centering
    \setlength{\tabcolsep}{0.5mm}
    \begin{tabular}{c|ccc|c}
    \toprule[1pt]
         Model & \makecell{Throughput\\(img/s)} & \makecell{Params\\(M)} & \makecell{FLOPs\\(G)} & \makecell{Acc\\(\%)} \\
         \midrule
         Swin-T & 1704 & 29 & 4.5 & 81.3 \\
         \midrule
         RMT-Swin-T & 1192 & 29 & 4.7 & 83.6 \\
         EVT-Swin-T & 1420 & 29 & 4.8 & 83.9 \\
         RetNet-Swin-T-linear & \textbf{336} & 29 & 4.3 & \textbf{80.8} \\         
    \bottomrule[1pt]
    \end{tabular}
    \caption{Comparison between RMT/EVT and RetNet.}
    \label{tab:comprer}
\end{table}

We compare the performance of various models on image classification, object detection, instance segmentation, and semantic segmentation tasks in detail. The results are shown in Tab.~\ref{tab:baseline}. EVT achieves excellent results across various downstream tasks. Specifically, in the image classification task, EVT-Swin-S achieves a top-1 accuracy of \textbf{85.1}, an improvement of \textbf{0.6} over RMT-Swin-S, and its inference speed is also somewhat faster. EVT-Swin-B achieves a top-1 accuracy of \textbf{85.3} with minimal speed loss, surpassing many deeper and larger models. In dense prediction downstream tasks, EVT-Swin also performs exceptionally well. For example, EVT-Swin-S surpasses RMT-Swin-S by \textbf{1.0} in $AP^b$, leads by \textbf{0.9} in $AP^m$, and shows an improvement of \textbf{0.7} in mIoU.

\noindent\textbf{Local Context Enhance Module. }The LCE module is a simple DWConv that enhances the model's ability to express local details, thereby improving its performance. As shown in Tab.~\ref{tab:baseline}, LCE improves the model's classification accuracy by \textbf{0.2} and significantly boosts performance in dense prediction tasks such as instance segmentation and semantic segmentation. Specifically, LCE provides the model with a performance increase of \textbf{+0.7}$AP^b$, \textbf{+0.6}$AP^m$, and \textbf{+0.7}mIoU.

\noindent\textbf{Conditional Position Encoding. }CPE is a positional encoding used in many models~\cite{stvit, CPVT, fan2024semantic, fan2024vision, biformer, FAT, cloformer}. It is a plug-and-play module composed of a simple DWConv, which conveniently provides positional information to the model. As shown in the Tab.~\ref{tab:baseline}, in EVT, CPE brings a slight performance improvement. Specifically, CPE enhances the model by \textbf{+0.1}top1-acc, \textbf{+0.3}$AP^b$, \textbf{+0.2}$AP^m$, and \textbf{+0.3}mIoU.

\noindent\textbf{Convolution Stem. }Convolution Stem helps to train the model more stably. Compared to patch embedding, Convolution Stem can extract better shallow fine-grained features, thereby enhancing the model's performance across various vision tasks. As shown in Tab.~\ref{tab:baseline}, in EVT, Convolution Stem improves the model by \textbf{+0.2}top1-acc, \textbf{+0.5}$AP^b$, \textbf{+0.7}$AP^m$, and \textbf{+1.2}mIoU.

\noindent\textbf{${\rm EuSA_g}$ and ${\rm EuSA_d}$. }${\rm EuSA_g}$ and ${\rm EuSA_d}$ are two one-dimensional token grouping methods we designed, which do not consider spatial information. They model short-range and long-range dependencies between tokens, respectively. In the Methods section, we discovered that under the influence of the Euclidean distance decay matrix, the one-dimensional grouping method outperforms the two-dimensional grouping method and is more flexible. In Tab.~\ref{tab:baseline}, we compare the effects of ${\rm EuSA_g}$ and ${\rm EuSA_d}$ on the model's performance. We found that using only one grouping method significantly harms the model's performance, which fully demonstrates the rationality of the two grouping designs.

\begin{table*}[t]
    \centering
    \begin{tabular}{c|c|cc|cccc}
    \toprule[1pt]
        Model & \makecell{Throughput\\(imgs/s)} & Parmas(M) & FLOPs(G) & Acc(\%) & $AP^b$ & $AP^m$ & mIoU \\
        \midrule[0.5pt]
        Swin-T & 1704 & 29 & 4.5 & 81.3 & 43.7 & 39.8 & 44.5 \\
        +CPE & 1640 & 29 & 4.5 & 81.7(\textcolor{red}{+0.4}) & 44.2(\textcolor{red}{+0.5}) & 40.5(\textcolor{red}{+0.7}) & 45.2(\textcolor{red}{+0.7}) \\
        +LCE & 1529 & 29 & 4.5 & 82.0(\textcolor{red}{+0.7}) & 44.6(\textcolor{red}{+0.9}) & 41.1(\textcolor{red}{+1.3}) & 45.6\textcolor{red}{+1.1}) \\
        +1D group/dilated & 1446 & 29 & 4.8 & 83.1(\textcolor{red}{+1.8}) & 46.6(\textcolor{red}{+2.9}) & 43.1(\textcolor{red}{+3.3}) & 46.9\textcolor{red}{+2.4}) \\
        \makecell{+Euclidean distance\\(EVT-Swin-T)} & 1420 & 29 & 4.8 & 83.9(\textcolor{red}{+2.6}) & 48.6(\textcolor{red}{+4.9}) & 43.9(\textcolor{red}{+4.1}) & 49.6(\textcolor{red}{+5.1})\\
    \bottomrule[1pt]
    \end{tabular}
    \caption{Roadmap from Swin-T to EVT-Swin-T.}
    \label{tab:roadmap2}
\end{table*}

\begin{table}[t]
    \centering
    \begin{tabular}{cc|ccc}
    \toprule
        Attention & \makecell{FLOPs\\(G)} & \makecell{Spatial\\Prior} & Acc(\%) & \makecell{Throughput\\(imgs/s)}\\
        \midrule[0.5pt]
        CSwin~\cite{cswin} & 4.6 & $\times$ & 83.7 & 924\\
        CSwin~\cite{cswin} & 4.6 & $\checkmark$ & 84.0 & 901\\
        \midrule[0.5pt]
        CCnet~\cite{huang2018ccnet} & 4.8 & $\times$ & 83.0 & 678\\
        CCnet~\cite{huang2018ccnet} & 4.8 & $\checkmark$ & 83.5 & 662 \\
        \midrule
        1D,98tokens & 4.6 & $\times$ & 83.6 & 1022\\
        1D,98tokens & 4.6 & $\checkmark$ & 84.4 &1001\\
        1D,49tokens & 4.4 & $\times$ & 83.5 & 1082\\
        1D,49tokens & 4.4 & $\checkmark$ & 84.2 & 1057\\
        1D,32tokens & 4.3 & $\times$ & 83.5 & 1123\\
        1D,32tokens & 4.3 & $\checkmark$ & 84.1 & 1100\\
        \bottomrule[1pt]
    \end{tabular}
    \caption{Ablation study on different models. All experiments are conducted based on the EVT-S. Throughput is measured by A100 80G with the batch size of 64.}
    \label{tab:attn_comp}
\end{table}

\noindent\textbf{Euclidean distance-based decay matrix. }The Euclidean distance-based decay matrix is a core module in EuSA, providing the model with spatially related prior knowledge. In Tab.~\ref{tab:baseline}, we validate the effect of the decay matrix on the model, demonstrating a significant performance improvement. Specifically, the introduction of the Euclidean distance-based decay matrix boosts the model's performance by \textbf{+0.7}top1-acc, \textbf{+0.7}$AP^b$, \textbf{+0.7}$AP^m$, and \textbf{+1.7}mIoU. To further understand the working mechanism of the decay matrix, we experiment with different configurations of the decay coefficient in Tab.~\ref{tab:gamma}. We find that setting different decay coefficients for each head in multi-head attention allows each head to focus on different scales, resulting in better model performance. Conversely, using the same decay coefficient for all heads leads to only a slight performance improvement. This indicates that the multi-scale information introduced by the decay matrix is what truly enhances the model's performance.

\noindent\textbf{Roadmap from RMT to EVT. }As shown in Tab.~\ref{tab:roadmap}, we gradually modify components in RMT-T to transform it into EVT-T. First, we adopt a deeper model, resulting in performance improvements (top1-acc\textbf{+0.1}, mIoU\textbf{+0.3}). Next, we replace the Manhattan distance-based decay matrix with a Euclidean distance-based decay matrix. This results in a further improvement in the model's performance (top1-acc\textbf{+0.4}, mIoU\textbf{+1.2}). Finally, we replace the horizontal and vertical decomposition in MaSA with the token grouping method of EuSA. This change not only reduces the computational load (by 0.2G) but also improves the model's performance (top1-acc\textbf{+0.6}, mIoU\textbf{+1.9}).

\noindent\textbf{A Comparative Discussion on RMT/EVT and RetNet. }RMT/EVT is inspired by RetNet, aiming to transfer its potentially beneficial properties to ViT for vision tasks. However, the ability of RetNet to be unfolded into an RNN is not applicable to vision tasks. This is because the inherently bidirectional nature of vision is not well suited for representation as an RNN. In RMT/EVT, we use a 2D, bidirectional decay matrix to model the vision information, whereas in RetNet~\cite{retnet}, the decay matrix for sequence tasks is 1D and unidirectional. Without using the Softmax function, both can be expressed with the following formula:
$$X_{out}=(QK^T\odot D)V$$
In RetNet, $D$ is a diagonal matrix, which allows RetNet to be conveniently unfolded into the form of an RNN. In RetNet, D is a diagonal matrix, with all values above the diagonal set to zero, meaning that the current token can only associate with preceding tokens and has no relation to the following tokens. This allows RetNet to be conveniently unfolded into the form of an RNN.
However, in RMT/EVT, each position in $D$ is non-zero, meaning that the current token interacts simultaneously with both preceding and succeeding tokens. This creates a bidirectional modeling process, which fundamentally prevents the model from being unfolded into the form of an RNN, as RNNs are inherently designed for unidirectional sequence modeling.

To validate our perspective, we directly apply RetNet's 1D decay and its linear complexity RNN formulation to Swin-T. The results are shown in Tab.~\ref{tab:comprer}. Using the $D$ matrix and RNN formulation from RetNet effectively reduces the model's computational cost, achieving linear complexity. However, the 1D inference approach disrupts the inherent 2D nature of vision, and the RNN formulation compromises the parallelism of visual reasoning. As a result, both the model's performance and inference speed suffer a significant decline.

\noindent\textbf{Roadmap from Swin-T to EVT-Swin-T.} We show the roadmap from Swin-T to EVT-Swin-T in the Tab.~\ref{tab:roadmap2}. The results clearly demonstrate the different modules' effect.

\begin{table}[t]
    \centering
    \begin{tabular}{cc|cc}
    \toprule[1pt]
         Model & $n_{tokens}$ & FLOPs & mIoU \\
         \midrule[0.5pt]
         EVT-S & 448 & 936 & 49.8 \\
         EVT-S & 98 & 872 & 49.3 \\
         EVT-S & 49 & 864 & 49.1 \\
         \midrule[0.5pt]
         SViT-S~\cite{stvit} & -- & 926 & 48.6 \\
    \bottomrule[1pt]
    \end{tabular}
    \caption{Ablation study on different number of tokens based on the semantic segmentation.}
    \label{tab:highres}
\end{table}

\noindent{\textbf{Ablation study on the number of vision tokens.} In Tab.~\ref{tab:attn_comp}, we compare model performance under different token group sizes (such as groups of 49 or 32 tokens, with padding when necessary). Although our approach is not equivalent to CSWin under these settings, it consistently achieves better performance.

Even with smaller group sizes (such as 32 tokens per group), our model benefits from the inherent flexibility of the 1D grouping paradigm. Compared with other methods that use fixed attention patterns, the 1D paradigm is more flexible and diverse, enabling the model to learn richer features.}

For higher resolutions, we test various values of $n_{token}$. Thanks to the highly flexible nature of 1D grouping in 2D space, as the resolution increases, the token groupings become even more diverse. This strong variability enables the model to learn richer spatial patterns. As shown in the Tab.~\ref{tab:highres}. Even when the number of tokens per group decreases, the model still achieves strong performance with the help of the spatial prior.

\noindent{\textbf{Efficiency and Performance on Multiple Resoslution. }We show the model's resource consumption and performance in the Fig.~\ref{fig:resource}. EVT, like other models with linear complexity, exhibits a linear increase in resource requirements.
\begin{figure*}
    \centering
    \includegraphics[width=0.99\linewidth]{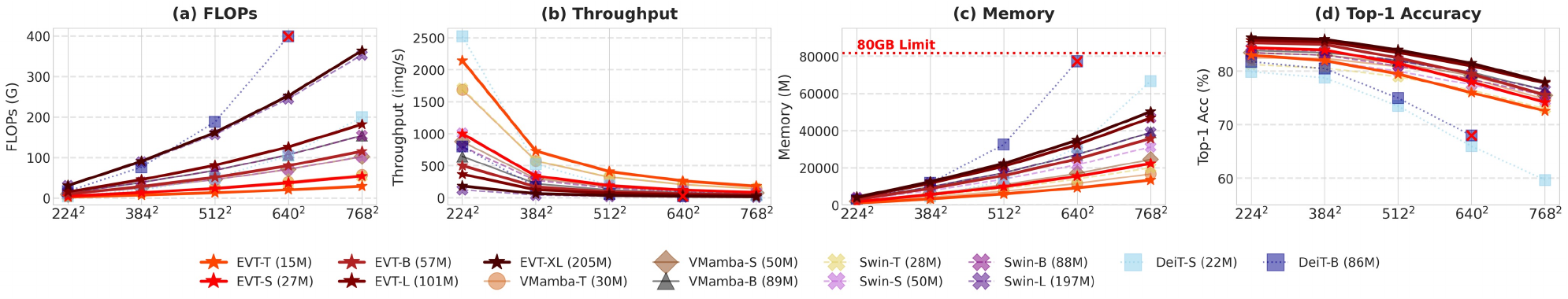}
    \caption{Illustration of EVT’s resource consumption and performance with progressively increasing resolutions}
    \label{fig:resource}
\end{figure*}}

\subsection{Visualization}

{\textbf{Euclidean distance v.s. Manhattan distance. }We show the visualization results in the Fig.~\ref{fig:thrattn}. The models are trained base on the DeiT. We can see that the incorporation of spatial prior makes the distribution of attention scores more concentrated, reducing the likelihood of attention dispersion. At the same time, compared to Manhattan distance, Euclidean distance brings smoother changes in attention, which more easily leads to a reasonable attention distribution.
\begin{figure}
    \centering
    \includegraphics[width=0.9\linewidth]{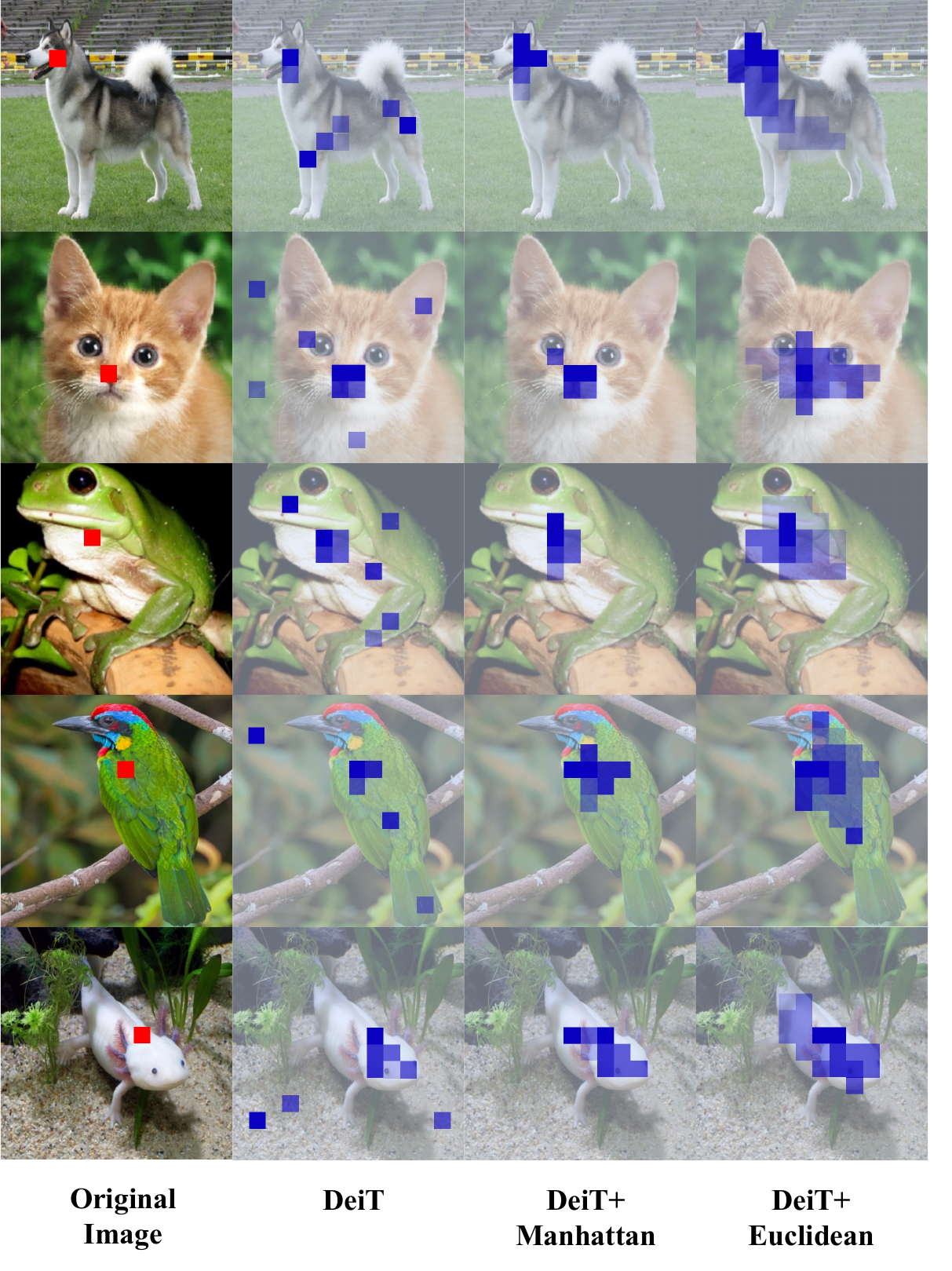}
    \caption{Visualization results of different models. Three models are DeiT-B (Top1-acc 81.8\%), DeiT-B+Manhattan distance (Top1-acc 82.4\%), DeiT-B+Euclidean (Top1-acc 82.7\%).}
    \label{fig:thrattn}
\end{figure}}

\begin{figure*}[!ht]
    \centering
    \includegraphics[width=0.74\linewidth]{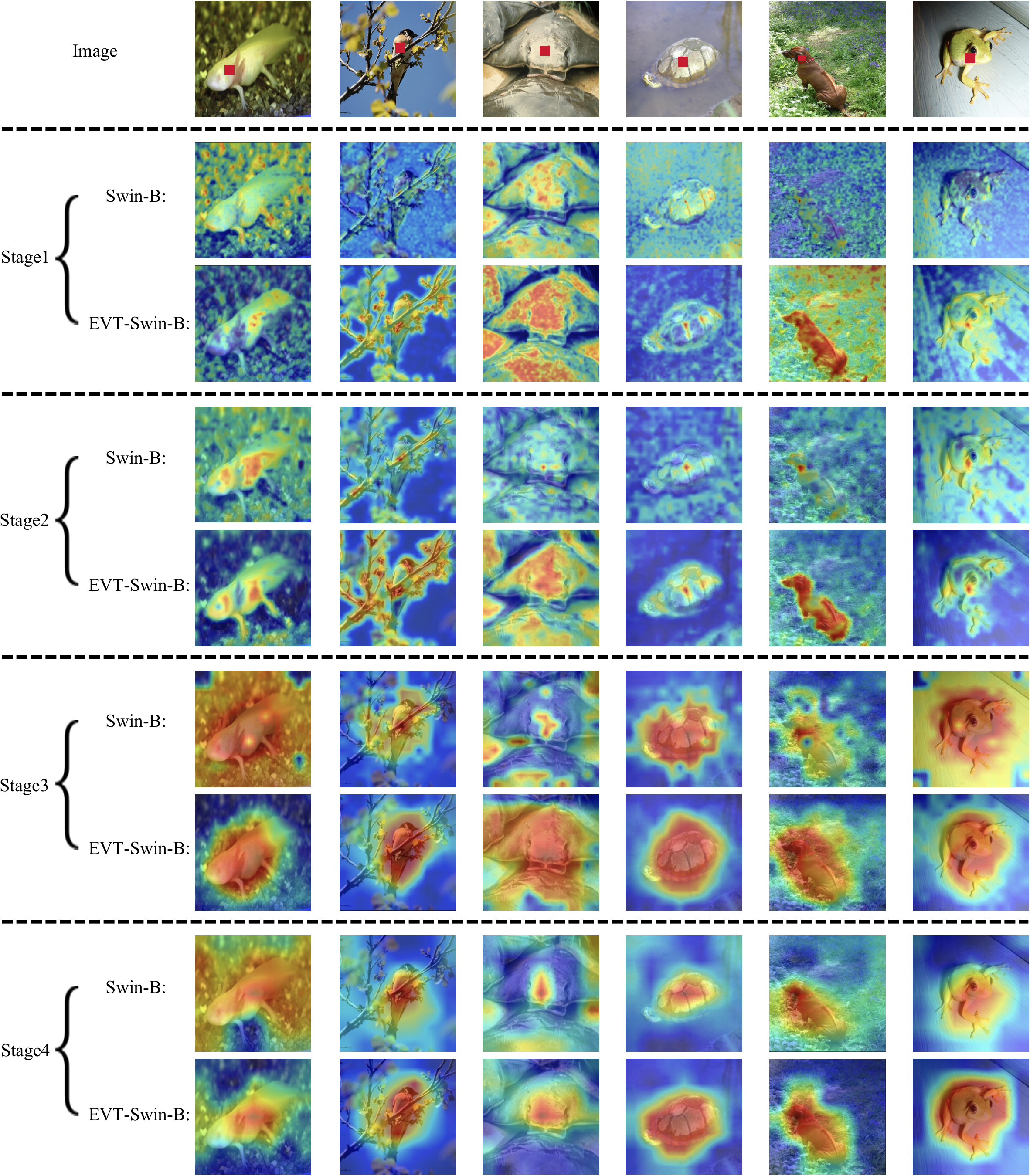}
    \caption{Visualization of EVT and Swin-Transformer's attention map.}
    \label{fig:visualization}
\end{figure*}

\noindent\textbf{Comparision with Swin. }To fully demonstrate the advantages of EVT, we visualize the attention maps of the tokens output at each stage. We also visualize the feature maps of Swin-Transformer for comparison. We visualize the attention maps at each hierarchical level of the Swin Transformer by constructing a global heat map for every stage. This approach is borrowed from the \textbf{Attention Rollout} technique used in \cite{attn_roll}, which aggregates attention scores across layers to create a global attention map.  To generate a global heat map at each level, we calculate attention scores for each query token by aggregating attention weights across all heads and groups, and we recursively accumulate these scores across layers. This leads to the formation of global attention patterns even from local attention operations. The results are shown in Fig.~\ref{fig:visualization}. We use an input image resolution of $224 \times 224$, and the token resolutions at Stage 1, 2, 3, and 4 are $56 \times 56$, $28 \times 28$, $14 \times 14$, and $7 \times 7$, respectively. From the visualization results, it is evident that EVT retains detailed information in the shallow layers of the model, such as Stage 1 and Stage 2. In the deeper layers, EVT successfully captures important object information in the images, enabling the model to achieve accurate classification. Compared to Swin-Transformer, EVT's feature map exhibits less noise and more accurate object localization, fully demonstrating the advantages of EVT.

\section{conclusion}
In this work, we propose EVT, a powerful general vision backbone. Building on RMT, it replaces the Manhattan distance-based explicit decay with a Euclidean distance-based explicit decay, resulting in improved model performance. Additionally, it replaces the $O(N^{1.5})$ complexity attention decomposition method in RMT with a more efficient one-dimensional token grouping approach of linear complexity. This makes EVT a powerful backbone that excels in both performance and efficiency. We validate the strong performance of the model in tasks such as image classification, object detection, instance segmentation, and semantic segmentation, and demonstrate its robustness on OOD data. Lastly, we conduct extensive ablation studies to verify the role of each module in the model.
\section{acknowledgements}
This work was supported by National Natural Science Foundation of China (Grant Nos. 62576342, 62425606, 62550062, 32341009), Beijing Natural Science Foundation (4252054, L257008), Beijing Nova Program (20230484276, 20240484601).

\bibliographystyle{IEEEtran}
\bibliography{ref}

\newpage
\section{Biography Section}

\begin{IEEEbiography}[{\includegraphics[width=1in,height=1.25in,clip,keepaspectratio]{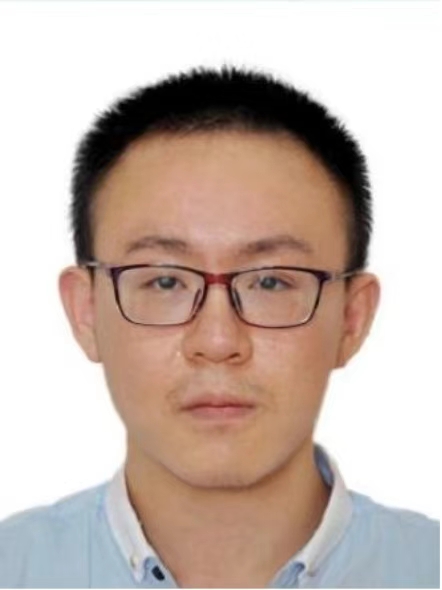}}]{Qihang Fan}
 received the B.Eng. degree with the Department of Precision Instrument, Tsinghua University, Beijing, China, in 2023. He is currently working toward the PhD degree with the Institute of Automation, Chinese Academy of Sciences, Beijing. His research interests include computer vision, deep learning architecture, and multimodal large language models.
\end{IEEEbiography}

\begin{IEEEbiography}[{\includegraphics[width=1in,height=1.25in,clip,keepaspectratio]{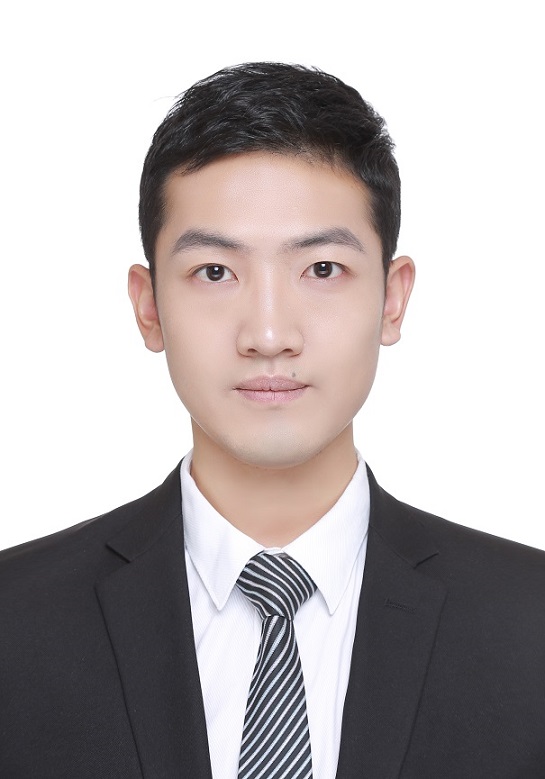}}]{Huaibo Huang}
 (Member, IEEE) received the BE degree in measurement and control technology and instrument from Xi’an Jiaotong University, in 2012, the ME degree in optical engineering from Beihang University, in 2016, and the PhD degree in pattern recognition and intelligent system from CASIA, in 2019. He is currently an associate professor in MAIS, NLPR, CASIA, China. His current research interests include computer vision and pattern recognition. He serves as the editor board member of IEEE TIFS and TBIOM.
\end{IEEEbiography}

\begin{IEEEbiography}[{\includegraphics[width=1in,height=1.25in,clip,keepaspectratio]{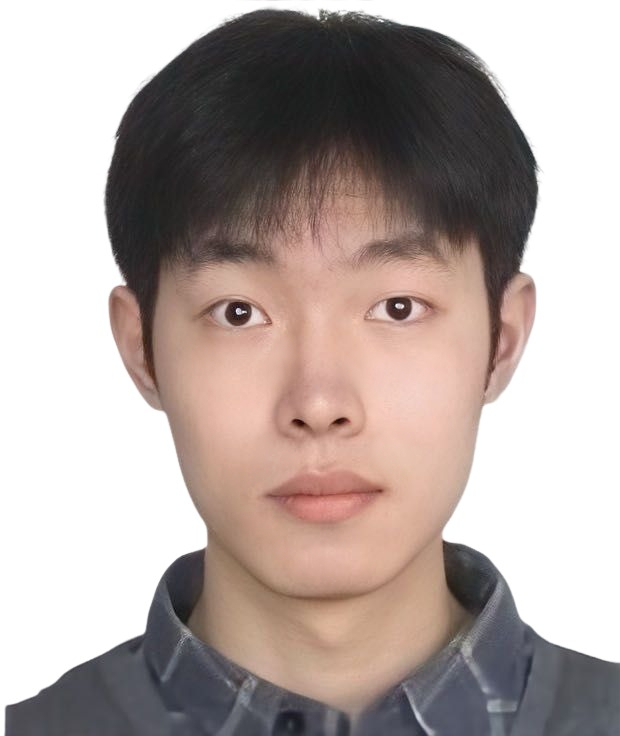}}]{Mingrui Chen}
 received the B.E. degree from the School of Artificial Intelligence and Automation, Huazhong University of Science and Technology, in 2024. He is currently working toward the Ph.D. degree with the School of Artificial Intelligence, University of Chinese Academy of Sciences (UCAS) and Institute of Automation, Chinese Academy of Sciences (CASIA), Beijing, China. His research interests include in deep learning architecture and representation learning.
\end{IEEEbiography}

\begin{IEEEbiography}[{\includegraphics[width=1in,height=1.25in,clip,keepaspectratio]{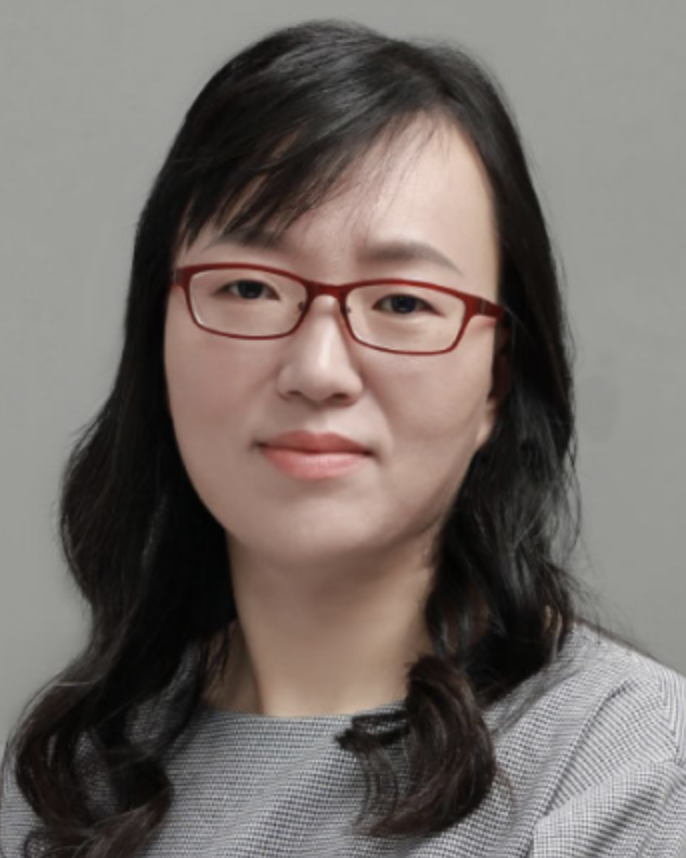}}]{Hongmin Liu}
received the B.S. degree from Xidian University, Xi’an, China, in 2004, and the Ph.D. degree from the Institute of Electronics, Chinese Academy of Sciences, Beijing, China, in 2009. She is currently a Professor with the School of Artificial Intelligence, University of Science and Technology Beijing. Her research interests include computer vision and smart sensing.
\end{IEEEbiography}

\begin{IEEEbiography}[{\includegraphics[width=1in,height=1.25in,clip,keepaspectratio]{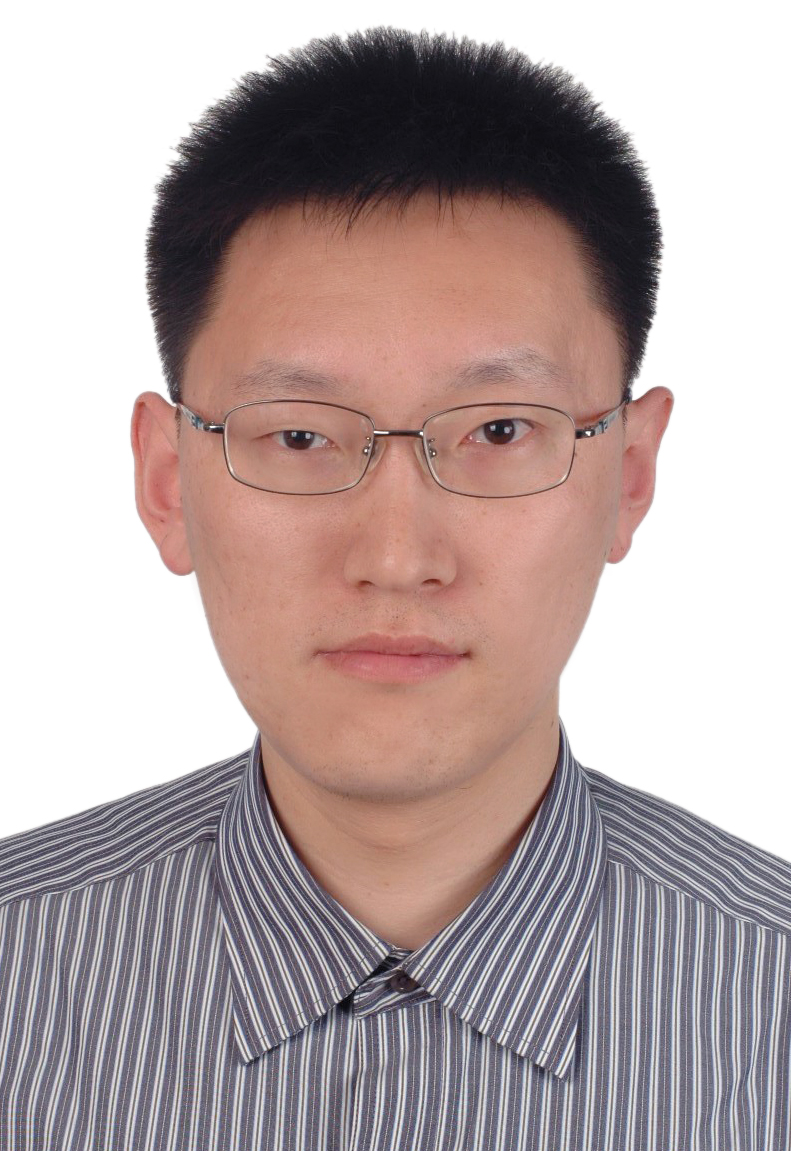}}]{Ran He}
 (Fellow, IEEE) received the BE degree in computer science from the Dalian University of Technology, in 2001, the MS degree in computer science from the Dalian University of Technology,
in 2004, and the PhD degree in pattern recognition and intelligent systems from CASIA, in 2009. Since September 2010, he joined NLPR, where he is currently a full professor. His research interests include information theoretic learning, pattern recognition, and computer vision. He serves as the editor board member of IEEE TPAMI, TIP, TIFS, TCSVT and TBIOM, and serves on the program committee of several conferences. He is also a fellow of the IAPR.
\end{IEEEbiography}

\end{document}